%% file: acl_latex.tex
\documentclass[11pt]{article}
\usepackage{amsmath}
\usepackage{amssymb}

\usepackage[preprint]{acl}

\usepackage{times}
\usepackage{latexsym}

\usepackage[T1]{fontenc}

\usepackage[utf8]{inputenc}

\usepackage{microtype}

\usepackage{inconsolata}

\usepackage{graphicx}
\usepackage{booktabs}
\usepackage{array}
\usepackage{multirow}
\usepackage[table]{xcolor}
\definecolor{skyblue}{RGB}{224,242,254}

\title{CIRF: Tokenizing Chain-of-Thoughts into Reusable Functional Units \\ for Efficient Latent Reasoning in Large Language Models}

\author{
 \textbf{Yukyung Lee\textsuperscript{1}}\;
 \textbf{Yumeng Shen\textsuperscript{2}}\;
 \textbf{Jinhyeong Park\textsuperscript{2}}\;
 \textbf{Hyein Yang\textsuperscript{2}}\;
 \textbf{Jun-Hyung Park\textsuperscript{2}}
\\
 \textsuperscript{1}Boston University\qquad
 \textsuperscript{2}Hankuk University of Foreign Studies
\\
 \texttt{ylee5@bu.edu}\qquad
 \texttt{\{yumengshen1023, asdjj, yhi, jhp\}@hufs.ac.kr}
}

\usepackage{longtable}
\begin{document}
\maketitle
\begin{abstract}
Implicit Chain-of-Thought (CoT) reduces the inference cost of large language models by internalizing the explicit rationales. However, existing approaches typically lack alignment with explicit rationales and adaptivity to example complexity. In this work, we propose CIRF (\textit{\underline{C}hain-of-thoughts \underline{I}nto \underline{R}eusable \underline{F}unctional units}), an implicit CoT framework that performs reasoning as a dynamic sequence of discrete functional tokens. CIRF assigns a functional token to each semantically coherent reasoning unit in explicit CoT traces. The model is then fine-tuned to autoregressively generate functional tokens and their optional results, followed by the final answer. This design aligns latent reasoning with a sequence of functional units, facilitating parallel training, explicit rationale alignment, and adaptive reasoning. Extensive experiments on mathematical, symbolic, and commonsense reasoning benchmarks show that CIRF provides a favorable accuracy-latency trade-off compared with state-of-the-art implicit CoT methods. Further analyses demonstrate that CIRF constructs distinct, interpretable functional tokens, leading to consistent performance improvements.

\end{abstract}

\begin{figure*}[!t]
\centering
\includegraphics[trim={0.25cm 12cm 7.25cm 0cm},width=\textwidth]{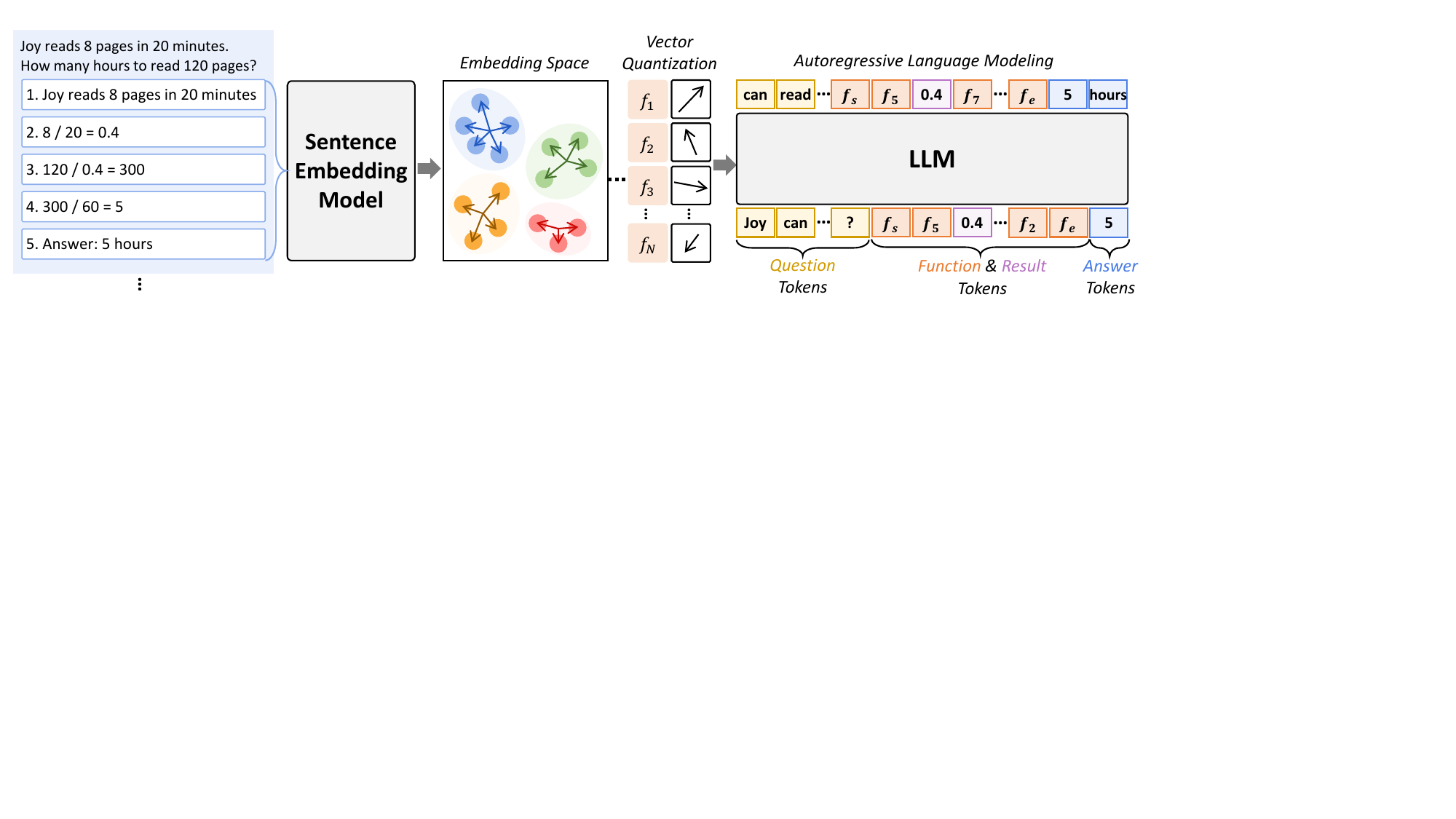}
\caption{Overview of CIRF. Each functional unit in a CoT rationale is encoded using a sentence embedding model, mean-centered to reduce question-specific situational bias, and quantized into a discrete functional token. The target LLM is then fine-tuned to generate a compact sequence consisting of functional tokens, optional result tokens, and the final answer tokens. CIRF enables efficient latent reasoning through discrete functional tokens.}
\label{figure_1} 
\end{figure*}

\section{Introduction}

Chain-of-thought (CoT) reasoning \cite{wei2022chainofthought,wang2023selfconsistency} has become a widely used strategy for eliciting multi-step reasoning in large language models (LLMs). By prompting or training a model to generate intermediate rationales before producing the final answer, CoT has substantially improved performance on mathematical and symbolic reasoning tasks. Recent analysis \cite{sprague2025tocot} suggests that the gains from CoT are strongest in domains that require symbolic or multi-step computation, further highlighting the importance of traceable intermediate steps for solving complex problems.

Despite its effectiveness, the latency and memory costs incurred by long reasoning traces have motivated research on implicit CoT \cite{deng2024explicittoimplicit,hao2025continuouslatent}. Instead of generating full rationales in natural language, these studies aim to internalize, compress, or replace explicit reasoning traces using hidden states or special tokens. They have shown that CoT can be performed implicitly within models, thereby reducing the cost of generating additional natural language tokens.

A key requirement for successful internalization is to align the LLM states during implicit CoT with those induced by explicit CoT. Typical implicit CoT methods utilize repeated placeholder tokens \cite{goyal2024pause,xu2025softcot} for alignment. However, these methods require the model to organize long, dynamic reasoning traces over a sequence whose surface form is largely homogeneous, whereas typical autoregressive LLMs are trained to reason conditioned on growing, changing textual prefixes. Whether such identical carriers provide an effective substrate for multi-step reasoning remains underexplored. 

To address this issue, several methods \cite{hao2025continuouslatent,shen2025codi} autoregressively feed output hidden representations as the next token embeddings, thereby reflecting changes in input token states as reasoning proceeds. However, training with recursively fed hidden representations incurs substantial training cost and instability, particularly in modeling multiple implicit tokens. In their formulations, input representations at each step need to be sequentially generated in training, because they are unknown until the model finishes computation at the previous step. Furthermore, their training procedures typically follow long curriculum stages or require a teacher model. The difficulty in scaling the number of implicit tokens may limit performance, considering that recent work \cite{hassid2025dontoverthink} increasingly shows that the number of reasoning tokens is important: too few tokens can lead to underthinking, while too many can cause overthinking, instability, or wasted computation. 

We propose a new implicit CoT framework, called CIRF (\textit{\underline{C}hain-of-thoughts \underline{I}nto \underline{R}eusable \underline{F}unctional units}), which represents reasoning as an autoregressively generated sequence of discrete functional tokens. The key idea is to tokenize explicit CoT traces into functional units. Consequently, each rationale is converted into a short, dynamic sequence of function-bearing tokens, enabling parallel training over sequences whose lengths adapt to the complexity of the reasoning process. We then fine-tune a single universal LLM to generate these implicit functional tokens, followed by optional results and the final answer for alignment, across diverse reasoning datasets. This design addresses the above issues by utilizing semantically grounded and adaptively allocated functional tokens.

Through extensive experiments, we empirically verify that CIRF is more efficient than state-of-the-art implicit CoT methods across diverse reasoning benchmarks. We evaluate CIRF in terms of accuracy-latency trade-off with diverse configurations partially mixing implicit and explicit tokens. The results show that CIRF achieves the accuracy–latency frontier. In-depth analyses further reveal that CIRF provides interpretable functional token assignment and adaptive token sequences, leading to substantial performance improvements.

\section{Related Work}
\subsection{Explicit CoT}
It has been observed that LLMs can generate intermediate reasoning steps that improve the accuracy of final answers \cite{wei2022chainofthought}. Zero-shot CoT \cite{kojima2022zeroshot} has shown that simple verbal triggers can elicit step-by-step reasoning without manually designed exemplars. While these findings highlight the importance of traceable intermediate steps for solving complex problems, many studies \cite{hassid2025dontoverthink,wu2026whenmore} have also reported an inverted-U relationship between CoT length and accuracy: reasoning traces that are either too short or too long can hurt performance. Several studies \cite{xu2025chainofdraft,xia2025tokenskip} have proposed methods for identifying shorter rationales that can match or exceed verbose CoT sequences. These studies are related to CIRF in that they show both the efficacy of generating intermediate reasoning steps and the importance of reasoning length for accuracy and efficiency.

\subsection{Implicit CoT}
Implicit CoT approaches aim to internalize reasoning into model hidden states. Early approaches have trained the hidden states associated with answer tokens to mimic the hidden states of explicit CoT \cite{deng2024explicittoimplicit}. Placeholder tokens \cite{goyal2024pause,pfau2024dotbydot} have also been used as learnable thinking symbols. However, these methods may limit the reasoning capabilities of LLMs due to the lack of semantically grounded tokens. Another line of work has conducted reasoning through recursively fed output hidden states \cite{hao2025continuouslatent,shen2025codi}. However, these approaches have introduced substantial complexity in training and offered limited adaptivity in reasoning length. A further subgroup of research has used auxiliary models to generate compact reasoning surrogates \cite{xu2025softcot,he2025semcot,wei2025simcot}, causing additional cost from the auxiliary models. Compared with these methods, CIRF provides a more efficient, straightforward strategy via a sequence of discrete functional tokens: facilitating single stage training with teacher forcing, realizing fine-grained reasoning functions and inducing adaptive reasoning behavior in a single unified model.
 
\subsection{Tokenization of Latent Representations}
\citet{wang2024planningtokens} have explicitly inferred planning variables using clustering over step representations, and its probing analysis has used pre-trained sentence encoders to study whether the induced planning categories are learnable and distinct. It has clustered reasoning steps to create auxiliary labels that precede explicit CoT. \citet{su2025tokenassorted} have used generated latent discrete tokens to randomly replace the initial portion of reasoning traces, showing that a discrete latent vocabulary can be learned. Compared with these methods, CIRF fully segments explicit CoT into semantically grounded units and introduces a carefully designed alignment method, leading to substantial improvements in reasoning efficiency.

\section{Methodology}
The proposed CIRF framework converts an explicit CoT rationale into a short sequence of discrete functional tokens and then fine-tunes a language model to autoregressively generate three components: discrete functional tokens, their corresponding results, and the final answer. In this section, we formulate the problem, and describe functional token assignment, alignment, and compression. 

\subsection{Problem Formulation}
We define a functional unit in CoT reasoning. Given a question \(x_i\), a typical LLM can generate an answer \(y_i\) together with a rationale \(r_i\) as follows:
\begin{equation}
(r_i, y_i) = G(x_i; p),
\end{equation}
where \(p\) is a prompt that elicits rationale generation. In standard CoT prompting, \(r_i\) is a reasoning trace composed of multiple textual reasoning steps. This trace can be represented as
\begin{equation}
r_i = (s_{i,1}, s_{i,2}, \ldots, s_{i,m_i}),
\end{equation}
where \(s_{i,j}\) denotes the textual description of the \(j\)-th reasoning step for question \(x_i\), and \(m_i\) is the total number of reasoning steps. We regard \(s_{i,j}\) as a description of a function that produces a new result \(u_{i,j}\), given contextual information including the question, prompt, and previous results \((x_i, p, u_{i,<j})\) as inputs. We then convert the textual output of explicit CoT into a sequence of functional tokens \(f_{i,j}\) and their corresponding result tokens \(u_{i,j}\) as follows:
\begin{equation}
(r_i, y_i) \rightarrow (f_{i,1}, u_{i,1}, f_{i,2}, u_{i,2}, \ldots, f_{i,m_i}, y_i).
\end{equation}
Our abstract formulation allows variable numbers of functional and result tokens for each reasoning step. We assign new tokens to the functions while using existing textual tokens for the results. Our goal is to construct a set of functional tokens whose latent representations are reusable across a wide range of questions, enabling an LLM to accurately induce the results through implicit reasoning.

\subsection{Functional Token Assignment}
We first segment \(r_i\) into semantically coherent reasoning units. In our preprocessing pipeline, segmentation is primarily delimiter-based: we detect explicit ordinal markers such as ``Step \(k\),'' ``\(k.\),'' or equivalent numbered boundaries. We treat the text span between two consecutive boundaries as one segment. For each reasoning segment \(s_{i,j}\), we compute a semantic representation using a pre-trained sentence embedding model \(G_{\mathrm{sent}}\):
\begin{equation}
z_{i,j} = G_{\mathrm{sent}}(s_{i,j}) \in \mathbb{R}^{d_s}.
\end{equation}
We use a sentence embedding model because such models are typically trained to produce semantically meaningful text representations suitable for similarity comparison, retrieval, and clustering. However, these representations are often biased toward the theme or situation of a given question, which makes it difficult to directly identify their functional types. Since the reasoning segments of a single question typically share a common situational bias, we reduce this bias by mean-centering the segment representations for each question:
\begin{equation}
z'_{i,j} = z_{i,j} - \frac{1}{m_i} \sum^{m_i}_{k=1} z_{i,k}.
\end{equation}
We empirically observe that this straightforward method effectively reduces situational bias while preserving functional information. We use the centered segment embeddings as the functional representations of reasoning segments.

We use VQ-VAE \cite{oord2017vqvae} to assign functional tokens to the segments. Let \(\mathcal{Z}=\{x_n\}_{n=1}^{M}\) denote the collection of all encoded functional representations in the training split, where \(x_n = F_{\mathrm{enc}}(z'_{i,j}) \in \mathbb{R}^{d_e}\) and \(M=\sum_i m_i\). Note that we set \(d_e\) to the target LLM embedding dimension and use a two-layer feedforward network with \(\textrm{tanh}\) activation for \(F_{\mathrm{enc}}(\cdot)\). We then initialize the codebook for vector quantization using a balanced clustering procedure based on the Sinkhorn-Knopp algorithm \cite{knight2008sinkhorn}. Specifically, we first choose \(K\) provisional anchors \(\{\tilde e_k\}_{k=1}^{K}\) from \(\mathcal{Z}\), and construct a positive affinity matrix:
\begin{equation}
A \in \mathbb{R}_{>0}^{M \times K},
A_{n,k} = \exp\!\left(-\frac{\lVert x_n-\tilde e_k\rVert_2^2}{\lambda}\right),
\end{equation}
where \(\lambda>0\) is a temperature parameter that controls the sharpness of the affinity distribution. We then alternately rescale the rows and columns of \(\mathbf{A}\) until the resulting matrix \(\mathbf{Q}\in\mathbb{R}_{>0}^{M\times K}\) lies on the doubly stochastic polytope:
\begin{equation}
\textstyle\sum_{k}Q_{n,k}=1,\qquad \sum_{n}Q_{n,k}=M/K.
\end{equation}
This procedure ensures that each anchor receives an equal share of assignment mass by construction. After obtaining the balanced soft assignment matrix, we convert it into a hard code assignment:
\begin{equation}
a_n = \arg\max_{k \in \{1,\ldots,K\}} Q_{n,k}.
\end{equation}
Equivalently, for each segment \(s_{i,j}\), we write
\begin{equation}
a_{i,j} = \arg\max_{k \in \{1,\ldots,K\}} Q_{(i,j),k},
\end{equation}
where \(Q_{(i,j),k}\) denotes the assignment probability of segment \(s_{i,j}\) to code \(k\). Let \(q_{i,j} = e_{a_{i,j}}\) denote the selected code vector for segment \(s_{i,j}\). The codebook is then learned using the VQ-VAE objective:
\[
\mathcal{L}_{\mathrm{vq}}
=
\frac{1}{\sum_i m_i}
\sum_{i=1}^{N}\sum_{j=1}^{m_i}
\Big[
\|F_{\mathrm{dec}}(q_{i,j}) - z'_{i,j}\|_2^2 +\qquad
\]
\begin{equation}
\qquad
\|\operatorname{sg}[x_{i,j}] - q_{i,j}\|_2^2 +
\beta \|x_{i,j} - \operatorname{sg}[q_{i,j}]\|_2^2
\Big],
\end{equation}
where \(F_{\mathrm{dec}}\) is another feedforward network, \(\operatorname{sg}[\cdot]\) is the stop-gradient operator, and \(\beta\) is the commitment coefficient. We use the same architecture for \(F_{\mathrm{enc}}\) and \(F_{\mathrm{dec}}\). The first term is a reconstruction error, which optimizes \(F_{\mathrm{dec}}\) to reconstruct the original centered representations from the functional code vectors. The second term updates the codebook vectors toward their assigned encoded representations, while the third term updates \(F_{\mathrm{enc}}\) to encourage the encoded representations to remain close to their quantized code vectors. We iteratively update the codebook vectors and reassign the codes according to the procedures described above.

After quantization, each segment \(s_{i,j}\) is replaced by a discrete functional token \(\tau_{a_{i,j}}\). Thus, a rationale \(r_i\) can be converted into a functional token sequence \((\tau_{a_{i,1}}, \tau_{a_{i,2}}, \ldots, \tau_{a_{i,m_i}})\).
To integrate these codes into the target LLM, we extend its vocabulary with \(K\) new special tokens \(\{\tau_1, \ldots, \tau_K\}\), together with dedicated tokens indicating the start and end of the reasoning span \(\tau_{\mathrm{s}}, \tau_{\mathrm{e}}\). 

Finally, we initialize the new token embeddings using the rescaled codebook vectors to transfer their semantic and geometry structure: 
\begin{equation}
    e_{k} \leftarrow \alpha \cdot \frac{e_{k}}{\lVert e_{k}\rVert_2},
\end{equation}
where \(e_{k}\) denotes a codebook vector and \(\alpha\) is a scaling parameter fixed to 0.01. This initialization allows the language model to exploit geometric relationships among functional codes from the beginning of training, leading to a better initial understanding of their functional roles.

\subsection{Result Collection and Alignment}

We introduce optional result tokens in latent reasoning, which denote the output of functional operations. The purpose of these result tokens is to provide the model with a minimal textual bridge between latent reasoning and answer generation. For each training example, we construct a result target \(u_i\) by prompting an LLM under three constraints: (1) the model should generate the result of a reasoning segment, rather than its intermediate derivation; (2) the model should generate only the result used in subsequent reasoning steps; and (3) the model should describe the result as concisely as possible. We denote this process as
\begin{equation}
u_i = G(x_i, r_i, y_i; p_{\mathrm{result}}),
\end{equation}
where \(p_{\mathrm{result}}\) is a prompt that instructs the model to produce the results. The result \(u_i\) can be decomposed into step result \((u_{i,1},u_{i,2},\ldots,u_{i,m_i})\) and empty step results are allowed. 

We aim to minimize textual tokens used for the results for two reasons. First, these tokens incur additional generation cost, which degrades inference efficiency. Second, we observe that exposing models to natural language reasoning traces can cause the latent token representations to collapse, as the models tend to favor reasoning in natural language over reasoning with unseen functional tokens.

For each example, the final supervision target is a token sequence:
\[
T_i = [\tau_{\mathrm{sof}}] \oplus \mathcal{H}(F_{i,1},...,F_{i,m_i}) \oplus [\tau_{\mathrm{eof}}, y_i]
\]
\begin{equation}
    s.t. \quad F_{i,j}=(\tau_{a_{i,j}}, u_{i,j})
\end{equation}
Here, \(\oplus\) denotes concatenation and \(\mathcal{H}\) is a token sequence construction function. The target LLM \(M_{\theta}\) is then fine-tuned using the standard autoregressive language modeling objective:
\begin{equation}
\mathcal{L}_{\mathrm{LM}}
=
-\sum_{i=1}^{N} \sum_{t=1}^{|T_i|}
\log p_{\theta}(T_{i,t} \mid x_i, T_{i,<t}).
\end{equation}

At inference time, CIRF requires only the fine-tuned target model \(M_{\theta}\). Given a question, the model generates the functional tokens and their optional result tokens, and subsequently predicts the final answer. 

\subsection{Result Compression}

In many examples (>70\%), we observe that functional token sequences alone are sufficient to generate correct answers. We therefore introduce a result compression procedure that retains only loss-reducing result units. Given the full set of result units \(U_i\), we construct a compressed result set \(C_i\) through iterative pruning. Starting from \(C_i^{(0)}=U_i\), we use a trained CIRF model as a scoring model to estimate the usefulness of each candidate. 

Given \(C_i^{(r)}\) at iteration \(r\), we greedily identify the candidate unit whose removal results in the smallest increase in loss. If the best candidate increases the loss by more than a predefined threshold \(\gamma\), we terminate the search. This greedy pruning strategy provides an explicit mechanism for controlling the accuracy-latency trade-off through the threshold \(\gamma\): a larger \(\gamma\) yields shorter tokens with lower decoding cost, whereas a smaller \(\gamma\) retains more result units when they improve answer likelihood.

\input{table/result_1}

\section{Experiments}
\label{sec:experiments}

In this section, we evaluate and compare CIRF with state-of-the-art efficient reasoning methods to demonstrate that it achieves an optimal accuracy-latency trade-off in latent reasoning.

\subsection{Experimental Setup}
\paragraph{Tasks and datasets.}
We evaluate CIRF on mathematical, symbolic, logical, and textual reasoning benchmarks. The mathematical benchmarks include GSM8K \cite{cobbe2021gsm8k}, SVAMP \cite{patel2021svamp}, MultiArith \cite{roy2015solving}, and MATH-500 \cite{hendrycks2021measuring}. The symbolic and logical benchmarks include Coin Flip \cite{wei2022chainofthought} and BIG-Bench Hard \cite{suzgun2023bbh}. The textual reasoning benchmarks include CommonsenseQA \cite{talmor2019commonsenseqa}, StrategyQA \cite{geva2021strategyqa}, and ScienceQA \cite{lu2022scienceqa}. GSM8K, SVAMP, MultiArith, Coin Flip, and CommonsenseQA are used as in-domain tasks for supervised fine-tuning and evaluation, whereas MATH-500, BIG-Bench Hard, StrategyQA, and ScienceQA are used as out-of-domain benchmarks. We provide further details on the evaluation protocol in Appendix \ref{datasets}.

\paragraph{Baselines.}
We compare CIRF with four categories of baselines. First, Direct Answer prompts the model to generate only the final answer, whereas SFT-CoT fine-tunes the model to generate the full explicit rationale followed by the answer. Second, the compressed explicit CoT baseline, i.e., Pause Tokens \cite{goyal2024pause}, reduces visible reasoning cost while remaining in natural language or repeated token space. Third, implicit CoT baselines, including iCoT-SI \cite{deng2024explicittoimplicit}, Coconut \cite{hao2025continuouslatent}, CODI \cite{shen2025codi}, and SIM-CoT \cite{wei2025simcot}, internalize reasoning through hidden states, continuous thought representations, or step-level latent supervision. Finally, soft token baselines, including SoftCoT \cite{xu2025softcot} and SemCoT \cite{he2025semcot}, generate compact reasoning surrogates using additional modules. We implement, train, and evaluate all baselines under controlled settings. The baseline settings are described in Appendix \ref{base_settings}.

\subsection{Main Results}

Figure~\ref{fig:main_tradeoff} compares accuracy and inference time across in-domain and out-of-domain settings. CIRF consistently lies in a favorable region of the Pareto frontier. CIRF$_{\textrm{Full}}$ achieves the highest accuracy among the CIRF variants, while CIRF$_{\textrm{Fast}}$ and CIRF$_{\textrm{Faster}}$ substantially reduce inference time by compressing result units. 

Specifically, with Qwen3-8B, all CIRF variants outperform the state-of-the-art baseline CODI in terms of accuracy. With a small additional latency, CIRF$_{\textrm{Full}}$ achieves higher accuracy than CODI, with a gap of up to 45.7\%p in GSM8K (CIRF$_{\textrm{Full}}$ 71.2\% vs. CODI 25.5\%). CIRF$_{\textrm{Fast}}$ and CIRF$_{\textrm{Faster}}$ exhibit better accuracy and latency than CODI, while CIRF$_{\textrm{Faster}}$ provides stronger acceleration than CIRF$_{\textrm{Fast}}$ at the cost of accuracy. In our experiments, compressing result units causes the largest accuracy degradation on Coin Flip (-20.9\%p for CIRF$_{\textrm{Fast}}$ and -20.7\%p for CIRF$_{\textrm{Faster}}$) and GSM8K (-6.5\%p for CIRF$_{\textrm{Fast}}$ and -16.5\%p for CIRF$_{\textrm{Faster}}$), which inherently require multiple reasoning steps and intermediate state tracking. Other baselines, such as Pause Tokens and soft token methods, show worse accuracy-latency trade-offs, possibly due to the limited reasoning capacity of a single placeholder token and the additional latency incurred by inference with multiple models.

We extend the Qwen3-8B experiments to out-of-domain tasks. CIRF achieves the most favorable accuracy-latency trade-offs, followed by CODI. CIRF consistently exhibits stable reasoning behavior, whereas most baselines produce degenerate results and long latencies on MATH-500, Big-Bench Hard, and StrategyQA, which also require multi-hop reasoning and state tracking. We further conduct experiments with another model, Llama3.1-8B, and observe consistent trends. CIRF achieves the Pareto frontier, with faster inferences at slightly lower accuracy. A notable difference in this experiment is the improvement of Pause Tokens, which suggests that Llama3.1-8B may better utilize placeholder tokens for latent reasoning. These results demonstrate the generalizability of CIRF across out-of-domain tasks and multiple models.

An interesting pattern is that all in-domain CIRF results except for CIRF$_{\textrm{Faster}}$ with Llama3.1-8B form a linear trend in the accuracy-latency plot. Based on this trend, we hypothesize that a scaling law may exist for latent reasoning with implicit tokens. We leave a more detailed examination of this pattern to future work.

\input{table/analyses}
\subsection{Analyses}
We conduct in-depth analyses to better understand the performance improvements. Unless otherwise specified, we use CIRF$_{\textrm{Full}}$ with Qwen3-8B.

\paragraph{Codebook size.}
In Figure \ref{fig:codebook_size}, we vary the number of functional tokens $K$ to study the granularity of the functional vocabulary. A very small codebook may merge distinct reasoning operations, whereas a very large codebook may fragment semantically similar operations and reduce token reuse. The codebook size ablation shows that moderate values of $K$ provide the best performance. Average performance gradually improves as the number of functional tokens increases, until it reaches an optimal value. Performance then decreases when the codebook becomes too large, suggesting that the functional vocabulary should be sized considering both the diversity and token frequency.

\paragraph{Adaptive reasoning.}
CIRF generates a variable-length sequence of functional tokens. As shown in Figure \ref{fig:length_difficulty}, we analyze whether this length reflects reasoning difficulty by grouping examples according to the number of generated functional tokens and measuring the corresponding error rate under direct answering. The results show a positive trend: examples requiring longer functional token sequences tend to have higher direct answering error rates. This trend indicates that CIRF performs latent reasoning adaptively with respect to instance difficulty and complexity, dynamically adjusting its reasoning length through the number of generated functional tokens.

\paragraph{Functional token interpretability.}
To examine whether functional tokens correspond to reusable reasoning operations, we retrieve reasoning segments assigned to frequent codes and inspect their shared semantics in Table \ref{tab:token_semantics}. The assigned segments reveal coherent operation types, such as answer selection, commonsense reasoning, addition, subtraction, and multiplication/division. This qualitative evidence suggests that the learned codes are neither arbitrary identifiers nor surface-level syntactic patterns, but instead capture reusable functional operations that recur across examples and datasets in an interpretable manner.

\begin{table}[t]
\centering
\small
\setlength{\tabcolsep}{4pt}
\begin{tabular}{lccc}
\toprule
Methods &
Bias $\downarrow$ &
Avg Cos. $\downarrow$ &
Max Cos. $\downarrow$ \\
\midrule
\multicolumn{4}{l}{\textit{Representation}} \\
\texttt{random}    & 0.128 & 0.000 & 0.109  \\
\texttt{raw}       & 0.405 & 0.164 & 0.955 \\
\texttt{q-cent.}   & 0.497 & 0.246 & 0.974 \\
\texttt{mean-cent.} & 0.124 & 0.009 & 0.756 \\
\midrule
\multicolumn{4}{l}{\textit{Codebook construction}} \\
K-means       & 0.131 & 0.011 & 0.869 \\
Sinkhorn      & 0.124 & 0.009 & 0.756 \\
\midrule
\multicolumn{4}{l}{\textit{Codebook size}} \\
$K=32$              & 0.124 & 0.009 & 0.756 \\
$K=128$              & 0.131 & 0.010 & 0.842 \\
$K=512$             & 0.110 & 0.010 & 0.919 \\
$K=2048$            & 0.085 & 0.007 & 0.939 \\
\bottomrule
\end{tabular}
\caption{
Functional token geometry under various settings. Bias denotes the bias share, defined as \(\|\mu\|/\mathbb{E}[\|w\|]\), where \(\mu\) is the mean vector of the LLM functional token embeddings and \(w\) is an individual token vector. Max/Avg Cos. denotes the maximum/average pairwise cosine similarity.
}
\label{tab:functional_token_geometry}
\end{table}

\paragraph{Functional token geometry.}
We further analyze the geometry of the functional token embeddings. This analysis examines whether the learned functional embeddings are geometrically diverse rather than collapsed into a single vector. Table \ref{tab:functional_token_geometry} shows four main observations. First, raw segment embeddings and question-centered embeddings exhibit large bias shares and high maximum cosine similarities in the final language model embedding space. This indicates that their token vectors are strongly affected by shared contextual offsets, which can reduce the separability of reusable functional operations. Second, the proposed Sinkhorn-initialized functional codebook maintains a low bias share, suggesting that balanced clustering effectively prevents collapse into a shared direction. Third, randomly initialized embeddings can appear geometrically well spread, as indicated by low cosine similarity and a low bias share, but such geometry is insufficient for functional reasoning because it does not preserve the semantic structure extracted from CoT segments. Finally, as the codebook size increases, the maximum cosine similarity also increases, indicating that overly large codebooks may introduce redundant functional tokens.

\section{Conclusion}
\label{sec:conclusion}

We presented CIRF, an implicit chain-of-thought framework that converts explicit reasoning traces into reusable discrete functional tokens. Through carefully designed functional token construction methods, CIRF generates dynamic sequences of functional tokens using a single autoregressive language model. Experiments across in-domain and out-of-domain reasoning benchmarks show that CIRF achieves a favorable accuracy-latency trade-off compared with state-of-the-art latent reasoning baselines, benefiting from its preservation of functional operational structure. We believe that CIRF offers a promising direction for building language models that reason more efficiently across a wide range of complex tasks.


\section*{Limitations}
\label{sec:limitations}

While we have demonstrated the frontier-level efficiency of CIRF, some limitations open promising avenues for future research. First, the quality of functional token supervision may depend on the quality and structure of the explicit CoT traces. We currently rely on explicit ordinal markers to segment rationales into reasoning units. More robust segmentation methods could further improve the reliability of the functional unit extraction process. In addition, the result token alignment and compression procedure introduces an additional design trade-off. An optimization method for this procedure may help identify a better accuracy-latency balance. Finally, as our experiments mainly focus on 8B-scale models and several tasks, the behavior of CIRF on differently sized models and additional tasks like long-context and tool-augmented reasoning would be explored in the future.

\section*{Ethics Statements}
CIRF is a general purpose framework for improving the efficiency of chain-of-thought reasoning in large language models. The method does not introduce new specific capabilities for harmful domains, nor does it rely on private, sensitive, or user-identifying data. The primary potential risk is indirect: because CIRF performs part of the reasoning process through discrete functional tokens rather than fully explicit natural-language rationales, it may reduce the human interpretability of individual reasoning steps compared with fully explicit CoT. However, CIRF partially mitigates this concern by constructing functional tokens from explicit CoT segments and by providing analyses showing that the learned tokens correspond to interpretable reasoning operations.



\bibliography{custom}

\appendix
\clearpage
\begin{table*}[t]
\centering
\small
\begin{tabular}{lcccccc}
\toprule
Dataset & Domain & Train & Test & Avg. CoT Tokens & Avg. Func. Tokens \\
\midrule
\textit{In-Domain Tasks}\\
GSM8K          & Math        & 7{,}473  & 1{,}319 & 194.8 & 3.95 \\
SVAMP          & Math        & 700      & 300     & 131.6 & 3.10 \\
MultiArith     & Math        & 420      & 180     & 130.3 & 3.56 \\
Coin Flip      & Symbolic    & 20{,}000 & 3{,}333 & 133.7 & 3.00 \\
CommonsenseQA  & Commonsense & 9{,}741  & 1{,}221 & 294.6 & 8.19 \\
\midrule
\textit{Out-of-Domain Tasks}\\
MATH-500        & Math       & -- & 500     & -- & -- \\
BIG-Bench Hard  & Symbolic   & -- & 6{,}511 & -- & -- \\
StrategyQA      & Multi-hop  & -- & 229     & -- & -- \\
ScienceQA       & Science    & -- & 4{,}241 & -- & -- \\
\bottomrule
\end{tabular}
\caption{
Dataset statistics for in-domain and out-of-domain evaluation.
In-domain tasks are used for supervised fine-tuning and evaluation, whereas out-of-domain tasks are used only for evaluation.
Average CoT tokens and average functional tokens are measured on Qwen3-8B.
}
\label{tab:dataset_statistics}
\end{table*}

\section{Detailed Experimental Settings}
\subsection{Tasks and Datasets}
\label{datasets}
\paragraph{Mathematical reasoning.}
We include multi-step arithmetic benchmarks, where explicit CoT has been consistently effective:
\begin{itemize}
    \item \textbf{GSM8K}~\cite{cobbe2021gsm8k}: grade-school math word problems requiring multi-step arithmetic reasoning.
    \item \textbf{SVAMP}~\cite{patel2021svamp}: adversarially constructed arithmetic word problems designed to test robustness to variations in problem structure.
    \item \textbf{MultiArith}~\cite{roy2015solving}: multi-step arithmetic problems with relatively short problem statements.
    \item \textbf{MATH-500}~\cite{hendrycks2021measuring}: competition-level mathematical reasoning problems. We use this dataset primarily as a harder out-of-distribution benchmark.
\end{itemize}

\paragraph{Symbolic and logical reasoning.}
To evaluate whether CIRF can represent reusable operations beyond arithmetic, we include symbolic and logical reasoning tasks:
\begin{itemize}
    \item \textbf{Coin Flip}~\cite{wei2022chainofthought}: a state-tracking task requiring updates over a sequence of operations.
    \item \textbf{BIG-Bench Hard}~\cite{suzgun2023bbh}: a wide range of tasks with each designed to evaluate different aspects of logical reasoning. 
\end{itemize}

\paragraph{Commonsense and multi-hop reasoning.}

To examine whether CIRF is useful on semantic reasoning steps, we include commonsense and multi-hop question answering datasets:

\begin{itemize}

    \item \textbf{CommonsenseQA}~\cite{talmor2019commonsenseqa}: multiple-choice commonsense reasoning.

    \item \textbf{StrategyQA}~\cite{geva2021strategyqa}: implicit multi-hop reasoning requiring decomposition into subquestions.

    \item \textbf{ScienceQA}~\cite{lu2022scienceqa}: science-domain multiple-choice reasoning. We use the text-only subset unless otherwise specified.
    
\end{itemize}

\subsection{CIRF Settings}
\label{cirf_settings}
\paragraph{CoT Collection}
We collect an explicit rationale using Gemma 3 27B \cite{gemmateam2025gemma3} and Llama 3.1 70B \cite{dubey2024llama3} on the training sets of GSM8K, SVAMP, MultiArith, CommonsenseQA, and Coin Flip for the base setting. We generate CoTs using a simple prompt of \textit{``Solve the following problem step by step:\{question\}''}. The rationale is normalized into a step-structured format before segmentation. We set the decoding temperature, top-p, and maximum generation length to 0.7, 0.95, and 4096, respectively. We filter out the entries that are not matched with our segmentation format, leading to elimination of 0.8\% of all generated entries.

\paragraph{Functional Token Construction}
We encode each segmented reasoning unit using the Qwen3-8B embedding model, and then construct a VQ codebook with the number of functional codes selected from $\{32, 64, 128, 256\}$ based on the validation accuracy, and the Sinkhorn temperature of 0.05. The dimension of code embedding vectors is identical to the embedding dimension of the reasoning model. We train the codebook for 10 epochs with the commitment coefficient of 1.0. The resulting code IDs are added to the backbone tokenizer as special tokens, and their embeddings are initialized by projecting the learned codebook vectors into the model embedding space.

\paragraph{Supervised Fine-Tuning}
Unless otherwise mentioned, we use AdamW with learning rate \(2\times 10^{-5}\), weight decay \(0.1\), and batch size \(32\). We train for 2 epochs and select the final checkpoint. We use cosine learning rate scheduling with 100 warmup iterations. We train models on NVIDIA RTX PRO 6000 GPUs with BF16 precision, taking approximately 1 GPU hour on the base settings.

\paragraph{Inference Settings}
We use nucleus sampling with top-p 0.95 and temperature 0.7. We provide one example and measure its correctness and latency, averaged over all examples. We execute three runs and report the average performance. For mathematical and symbolic reasoning tasks, we use exact-match accuracy after answer normalization. For multiple-choice tasks, we evaluate whether the predicted option matches the ground truth. We report end-to-end time from input to final answer completion as an efficiency metric.

\paragraph{Backbone models.}
We evaluate CIRF on multiple backbone families to verify that its effectiveness is not specific to a single model architecture. Unless otherwise stated, all backbone models are instruction-tuned causal language models, i.e., Llama3.1-8B and Qwen3-8B.

\paragraph{CIRF Variants}
We evaluate the following variants of CIRF to characterize the trade-off between reasoning quality and inference cost:
\begin{itemize}
    \item \textbf{CIRF}$_{\textrm{Full}}$: generates functional tokens and all aligned result units.
    \item \textbf{CIRF}$_{\textrm{Fast}}$: generates functional tokens and compressed result units selected by our loss-reduction criterion, whose threshold is set to 0.1.
    \item \textbf{CIRF}$_{\textrm{Faster}}$: generates functional tokens and more aggressively pruned result units with a threshold set to 0.2.
\end{itemize}

We list the hyper-parameter settings in Table \ref{tab:hyper_parameters}.

\subsection{Baselines}
\label{base_settings}
We compare CIRF with baselines from four categories. We implement the baselines based on their publicly released codes for scientific purpose, complying with their license and intended use. All baselines are trained on a single RTX PRO 6000 GPU with bfloat16 using the AdamW optimizer, unless otherwise noted. For the Math500 dataset, we set the maximum output length to $512$. Evaluation is performed at temperature $0.7$, averaged over three inference runs from a trained model. Most fine-tuned methods use LoRA~\citep{hu2022lora} with $r{=}128$ and $\alpha{=}32$, with Baseline-specific settings described below.

\paragraph{Direct and Explicit CoT Baselines}
These baselines represent the standard explicit CoT settings, which decode natural language rationales.
\begin{itemize}
    \item \textbf{Direct Answer}: the model is prompted to produce only the final answer. This baseline measures the performance of non-rationale generation.
    \item \textbf{SFT-CoT}: the backbone is supervised fine-tuned to generate the full explicit rationale followed by the final answer.
\end{itemize}

\paragraph{Compressed Explicit CoT Baselines}
These methods remain in natural-language token space. They are important baselines because CIRF also reduces visible reasoning length, but does so by replacing reasoning operations with discrete functional tokens.
\begin{itemize}
    \item \textbf{Chain-of-Draft (CoD)}~\cite{xu2025chainofdraft}: generates concise intermediate reasoning drafts instead of verbose CoT traces. Following the official implementation,\footnote{\url{https://github.com/xu3kev/chain-of-draft}} CoD is applied at inference time through prompting alone and requires no additional training.
    \item \textbf{TokenSkip}~\cite{xia2025tokenskip}: selectively skips less important reasoning tokens to reduce CoT length. We adopt the authors' implementation\footnote{\url{https://github.com/hemingkx/TokenSkip}} and fine-tune for $10$ epochs with a learning rate of $1\mathrm{e}{-}4$ and weight decay of $0.01$.
    \item \textbf{Pause Tokens}~\cite{goyal2024pause}: inserts repeated special tokens before answer generation to provide additional computation steps. We insert five learnable \texttt{<pause>} tokens before the answer and fine-tune for $10$ epochs with a learning rate of $1\mathrm{e}{-}4$ and weight decay of $0.01$, following the implementation released with SemCoT.\footnote{\url{https://github.com/YinhanHe123/SemCoT}}\label{fn:semcot}
\end{itemize}

\paragraph{Implicit CoT Baselines}
These methods are closest to CIRF in objective, as they aim to reduce or remove explicit natural-language rationales. However, they differ in whether the reasoning carrier is continuous, homogeneous, externally decoded, or semantically discrete.
\begin{itemize}
    \item \textbf{iCoT-SI}~\cite{deng2024explicittoimplicit}: distills explicit CoT supervision into hidden representations. We progressively remove the rationale during training, dropping eight tokens per epoch over $10$ epochs with a learning rate of $5\mathrm{e}{-}5$, following the implementation released with SemCoT.\footref{fn:semcot}
    \item \textbf{Coconut}~\cite{hao2025continuouslatent}: uses the previous hidden state as a continuous thought representation and feeds it back as the next input embedding. We use five continuous thought tokens learned through a three-stage curriculum and train for $10$ epochs with a learning rate of $1\mathrm{e}{-}5$ and weight decay of $0.01$, following the implementation released with SemCoT.\footref{fn:semcot}
    \item \textbf{CODI}~\cite{shen2025codi}: aligns explicit and implicit CoT through self-distillation. We adopt the official implementation\footnote{\url{https://github.com/zhenyi4/codi}, \footref{fn:semcot}} with six latent tokens, training for $5$ epochs with a learning rate of $8\mathrm{e}{-}4$, weight decay of $0.1$, a distillation factor of $\gamma{=}20$, an effective batch size of $128$, and a cosine schedule with warmup ratio $0.03$.
    \item \textbf{SIM-CoT}~\cite{wei2025simcot}: uses explicit step-level supervision to stabilize and diagnose latent reasoning states. Building on Coconut, we add auxiliary-decoder supervision with five thought tokens and train for $10$ epochs with a learning rate of $1\mathrm{e}{-}5$, weight decay of $0.01$, three curriculum stages, and an auxiliary loss weight of $1.0$, following the official implementation.\footnote{\url{https://github.com/InternLM/SIM-CoT}}
\end{itemize}

\paragraph{Auxiliary Model and Soft Token Baselines}
These baselines are essential for evaluating wall-clock efficiency, because auxiliary modules can reduce the number of decoded tokens but may introduce additional prompt-dependent computation.
\begin{itemize}
    \item \textbf{SoftCoT}~\cite{xu2025softcot}: uses a lightweight assistant model to generate instance-specific soft thought tokens for the main LLM. We train only the projection layer that maps the assistant's hidden states into five soft thought tokens, keeping both the assistant and the main model frozen, for $10$ epochs with a learning rate of $1\mathrm{e}{-}3$ and weight decay of $0.01$, following the implementation released with SemCoT.\footref{fn:semcot}
    \item \textbf{SemCoT}~\cite{he2025semcot}: introduces semantic alignment for implicit tokens to preserve reasoning information while accelerating generation. We follow the official implementation\footref{fn:semcot} and use five semantic thought tokens during training and a single token at evaluation.
\end{itemize}

\input{table/result_2}
\input{table/appendix_experiment}


\section{Additional results}
\label{full_results}

\paragraph{Functional representation extraction.}
We study how to extract functional information from reasoning-segment embeddings. We compare random embeddings (\texttt{random}), raw segment embeddings without centering (\texttt{raw}), segment embeddings centered by their corresponding question embedding (\texttt{q-cent.}), and the mean-centered embeddings (\texttt{mean-cent.}). As shown in Table \ref{tab:representation_accuracy}, \texttt{mean-cent.} achieves the best average downstream accuracy, improving over \texttt{raw} and \texttt{q-cent.}  These results support that mean-centering suppresses question-specific situational bias while preserving reusable reasoning functionality.

\paragraph{Codebook construction.}
We examine the effect of codebook initialization and token-embedding initialization in Table \ref{tab:codebook_init}. We compare random initialization, K-means initialization, Sinkhorn initialization, and methods without semantic embedding initialization. The results suggest that both the balanced code assignment and trained embeddings are important for learning reusable functional token embeddings during language model fine-tuning.

\paragraph{Codebook size analysis.}
We measure the delta accuracy, which is an accuracy difference from the minimum accuracy, for each codebook size, as shown in Figure \ref{fig:delta_codebook_size}.

\paragraph{Functional code analysis.}
We compare the clustering results of raw segment embeddings, question-centered embeddings, and the mean-centered embeddings. \textit{Used} denotes the fraction of activated codes. \textit{AMI} measures the dependence between assigned codes and question identity. \textit{Puri.} reports the size-weighted cluster purity with respect to question identity.\textit{ Coll.} denotes the fraction of examples whose segments are assigned to a single code. \textit{Uniq.} reports the mean number of distinct codes per example. We observe that the mean-centered representation substantially reduces the association between code assignments and question identity, as indicated by lower adjusted mutual information and lower cluster purity in Table \ref{tab:intrinsic_clustering}. These results support that mean-centering suppresses question-specific situational bias while preserving reusable reasoning functionality.

\paragraph{Full results.}
We report the task-wise accuracy and latency of the experimental results in Table~\ref{tab:main-results}.

\paragraph{Effect of model scale.}
We extend our evaluation to additional Qwen3 scales (1.7B, 4B, and 14B). As shown in Figure~\ref{fig:scaling_comparison}, the Pareto optimal CIRF variants dominate both state-of-the-art baselines (CODI and Pause) at every scale, showing that the accuracy and efficiency gains generalize across model sizes.

\paragraph{CIRF Variant Comparison.}
Figure~\ref{fig:cirf_comparison} compares CIRF variants aggregated across all backbone LLMs. The Fast and Faster variants substantially reduce latency while maintaining competitive accuracy. We observe that in-domain accuracies across different backbones remain similar with respect to inference time, whereas out-of-domain accuracies are different. A trend can be observed that stronger backbone models show stronger generalizability.

\section{Information About Use Of AI Assistants}
We have used AI assistants in language editing of the paper and implementation of partial code, all of which are double-checked by humans.

\input{table/hyperparams}
\input{table/main_experiment}
\end{document}

%% file: table/result_1.tex
\begin{figure*}[ht]
    \centering
    \includegraphics[width=\textwidth]{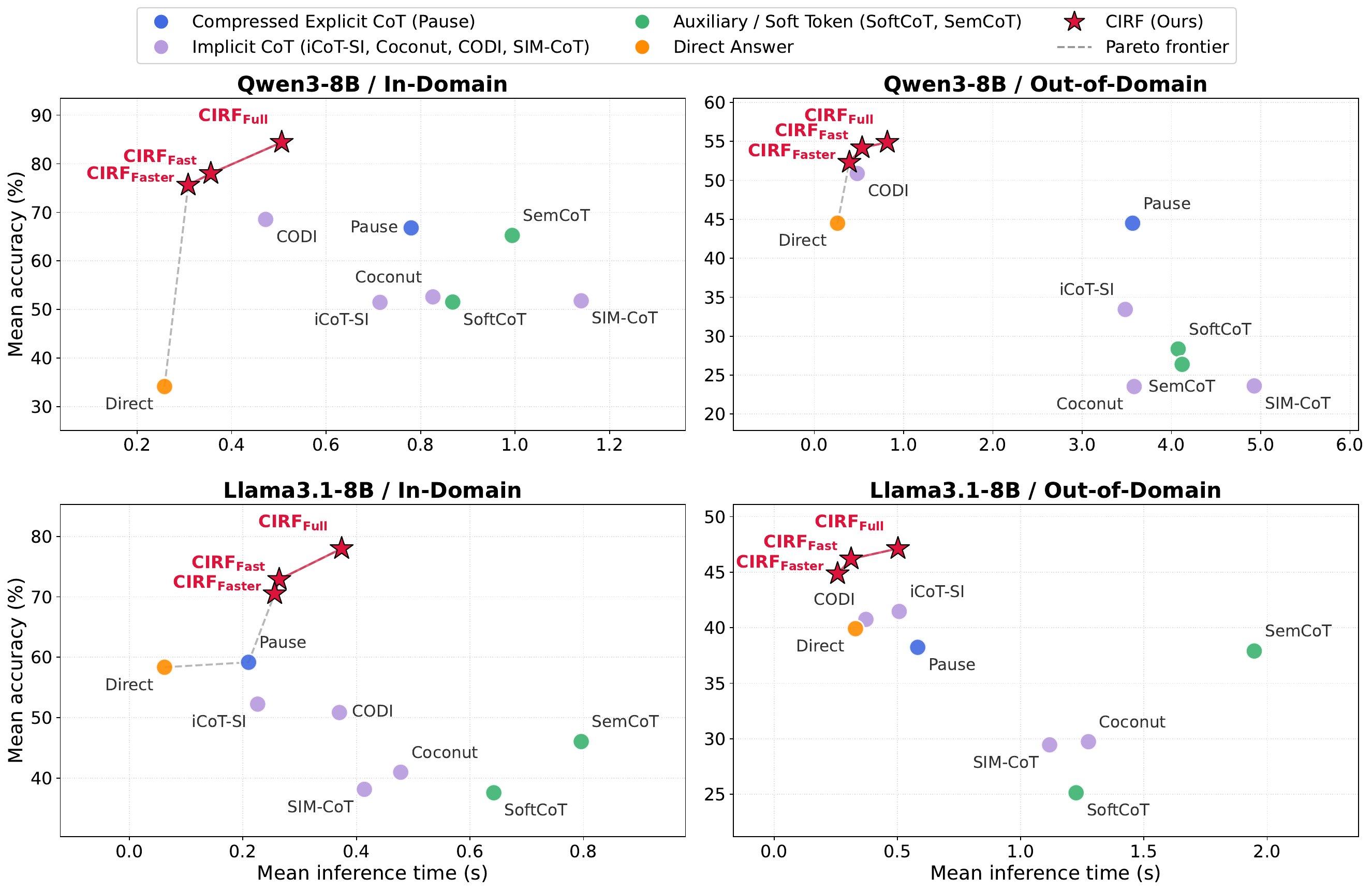}\caption{
Accuracy-latency trade-off across in-domain and out-of-domain benchmarks. Each point reports the mean accuracy and mean inference time of a reasoning method. Task-wise full results are reported in Appendix~\ref{full_results}.
}
\label{fig:main_tradeoff}
\end{figure*}

%% file: table/analyses.tex



\begin{figure*}[!t]
\centering
\includegraphics[width=\textwidth]{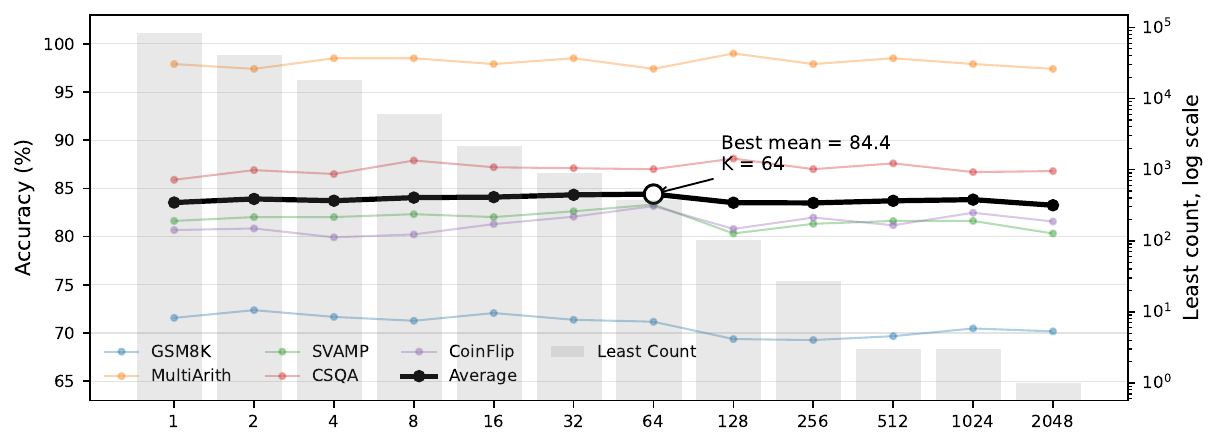}
\caption{
Effect of codebook size on downstream accuracy and post-training code utilization. The solid line reports average accuracy across benchmarks, the task-specific curves show per-dataset accuracy, and the gray bars indicate the count of the least used code on a logarithmic scale. Delta accuracy comparisons are reported in Appendix~\ref{full_results}.
}
\label{fig:codebook_size}
\end{figure*}


\begin{figure*}[!t]
\centering
\includegraphics[width= \textwidth]{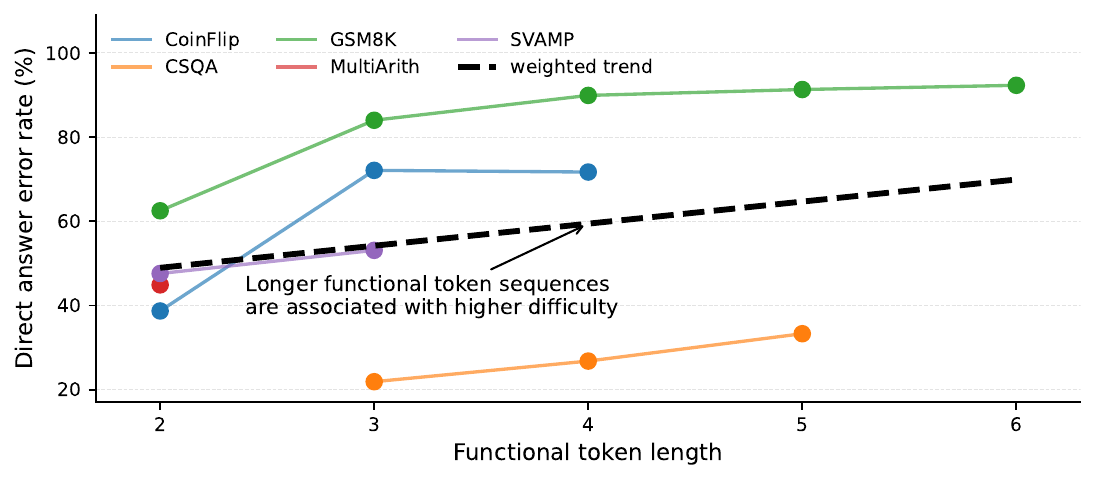}
\caption{
Relationship between generated functional token length and instance difficulty. Examples are grouped by the number of generated functional tokens, and the error rate is reported for each group.
}
\label{fig:length_difficulty}
\end{figure*}

\begin{table*}[t]
\centering
\small
\setlength{\tabcolsep}{4pt}
\begin{tabular}{lllll}
\toprule
Code & Frequency & Operation & Assigned CoT Segments \\
\midrule
25 & 3,528 & Choose answer & ... the best answer is (E) use television. This is because the use of a remote ... \\
&&& ... Based on the analysis, the most logical and relevant answer is (C) suffering pain, ... \\                           
&&& ... the best option is (C) dribble, as it accurately describes what happens to ... \\             
\midrule
18 & 3,335 & Commonsense & ... an attic is typically above the main living space of a house ... \\
&&& ... A beach is an outdoor location, often very bright due to the sun. ... \\                           
&&& ... Army captains might participate in recreational activities such as basketball, ... \\             
\midrule
23 & 3,051 & Mul/Div & 4 bananas * \$1/banana = \$4 \\
&&& ... She adds twice that amount: 3 statues * 2 = 6 statues. ... \\                           
&&& ... so she tried on 16 * 2 = 32 pairs. \\             
\midrule
16 & 2,745 & Addition & ... \$40 (coat) + \$30 (shoes) = \$70 \\
&&& ... Her total savings are \$50 + \$15 + \$30 = \$95. \\                           
&&& ... Each basket has 4 red peaches + 6 green peaches = 10 peaches. \\             
\midrule
3 & 2,635 & Subtraction & ... The wallet costs \$100, and Betty has \$95. She needs \$100 - \$95 = \$5 more ... \\
&&& ... Amount for insurance: \$5200 - \$4940 = \$260 ... \\                           
&&& ... 28 - 15 = 13 \\             
\bottomrule
\end{tabular}
\caption{
Qualitative examples of learned functional token assignments. For each code, we report its frequency, the interpreted operation type, and three CoT segments assigned to the code.
}
\label{tab:token_semantics}
\end{table*}

%% file: table/result_2.tex
\begin{table*}[t]
\centering
\small
\setlength{\tabcolsep}{4pt}
\begin{tabular}{lcccccc}
\toprule
Method & GSM8K & MultiArith & SVAMP & CommonsenseQA & CoinFlip & Avg. \\
\midrule
\texttt{random}               & 35.3 & 79.6 & 71.0 & 78.1 & 78.0 & 68.4 \\
\texttt{raw}               & 42.1 & 92.9 & 81.3 & 85.4 & 79.3 & 76.2 \\
\texttt{q-cent.}    & 69.0 & 96.8 & 82.7 & 87.3 & 81.8 & 83.5 \\
\texttt{mean-cent.}     & \textbf{71.2} & \textbf{97.4} & \textbf{83.3} & \textbf{87.0} & \textbf{83.2} & \textbf{84.4} \\
\bottomrule
\end{tabular}
\caption{
Downstream accuracy of functional representation methods after codebook training and CIRF fine-tuning.
}
\label{tab:representation_accuracy}
\end{table*}

\begin{table*}[t]
\centering
\small
\setlength{\tabcolsep}{4pt}
\begin{tabular}{llcccccc}
\toprule
Init. Code & Init. Emb. & GSM8K & MultiArith & SVAMP & CommonsenseQA & CoinFlip & Avg. \\
\midrule
Random      & Random  & 35.3 & 79.6 & 71.0 & 78.1 & 78.0 & 68.4 \\
Random      & Trained & \textbf{71.5} & 96.9 & 82.6 & \textbf{87.1} & 81.2 & 83.9 \\
K-means     & Trained & 71.0 & 96.4 & 80.6 & 86.7 & 81.9 & 83.3 \\
Sinkhorn    & Trained & 71.2 & \textbf{97.4} & \textbf{83.3} & 87.0 & \textbf{83.2} & \textbf{84.4} \\
Sinkhorn    & Random  & 70.8 & 95.2 & 81.3 & 87.1 & 79.0 & 82.7 \\
\bottomrule
\end{tabular}
\caption{
Codebook-construction ablation.
We compare different code initialization strategies and functional-token embedding initializations.
}
\label{tab:codebook_init}
\end{table*}

%% file: table/appendix_experiment.tex
\begin{figure*}[!t]
\centering
\includegraphics[width=\textwidth]{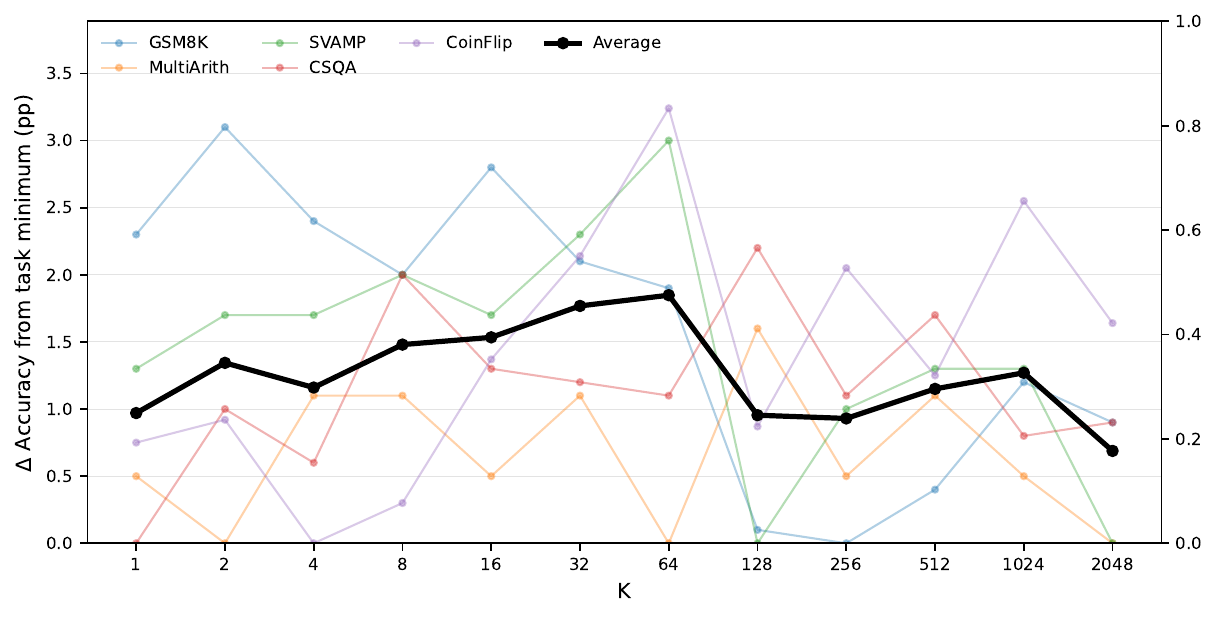}
\caption{
Effect of codebook size on downstream delta accuracy from task minimum and post-training code utilization.
The solid line reports average delta accuracy across benchmarks and task-specific curves show per-dataset accuracy.
}
\label{fig:delta_codebook_size}
\end{figure*}
\begin{table}[t]
\centering
\small
\setlength{\tabcolsep}{4pt}
\begin{tabular}{lcccccccccc}
\toprule
Method & Used $\uparrow$  & AMI $\downarrow$ & Puri. $\downarrow$ & Coll. $\downarrow$  & Uniq. $\uparrow$ \\
\midrule
\texttt{raw}                  & \textbf{1.000} & 0.113          & 0.010          & 0.070         & 3.28 \\
\texttt{q-cent.}    & \textbf{1.000} & 0.043          & 0.007          & 0.008          & 4.03 \\
\texttt{mean-cent.}        & \textbf{1.000} & \textbf{0.000}          & \textbf{0.003} & \textbf{0.000} & \textbf{4.55} \\
\bottomrule
\end{tabular}
\caption{
Intrinsic evaluation of functional representation methods for code assignment with $K=128$.
}
\label{tab:intrinsic_clustering}
\end{table}

\begin{figure*}[t]
    \centering
    \includegraphics[width=\linewidth]{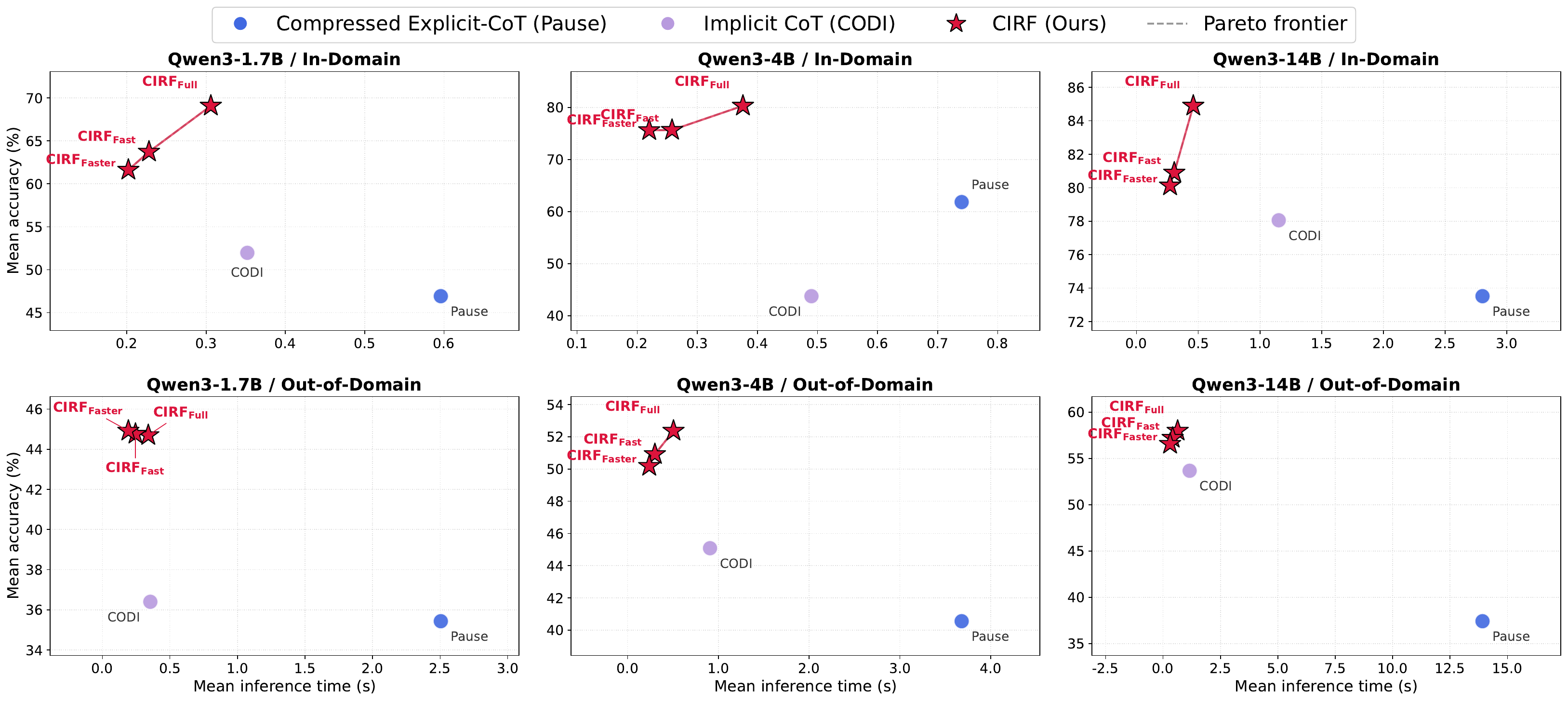}
    \caption{Scaling experiments on Qwen3-1.7B, 4B, and 14B.}
    \label{fig:scaling_comparison}
\end{figure*}

\begin{figure*}[t]
    \centering
    \includegraphics[width=\linewidth]{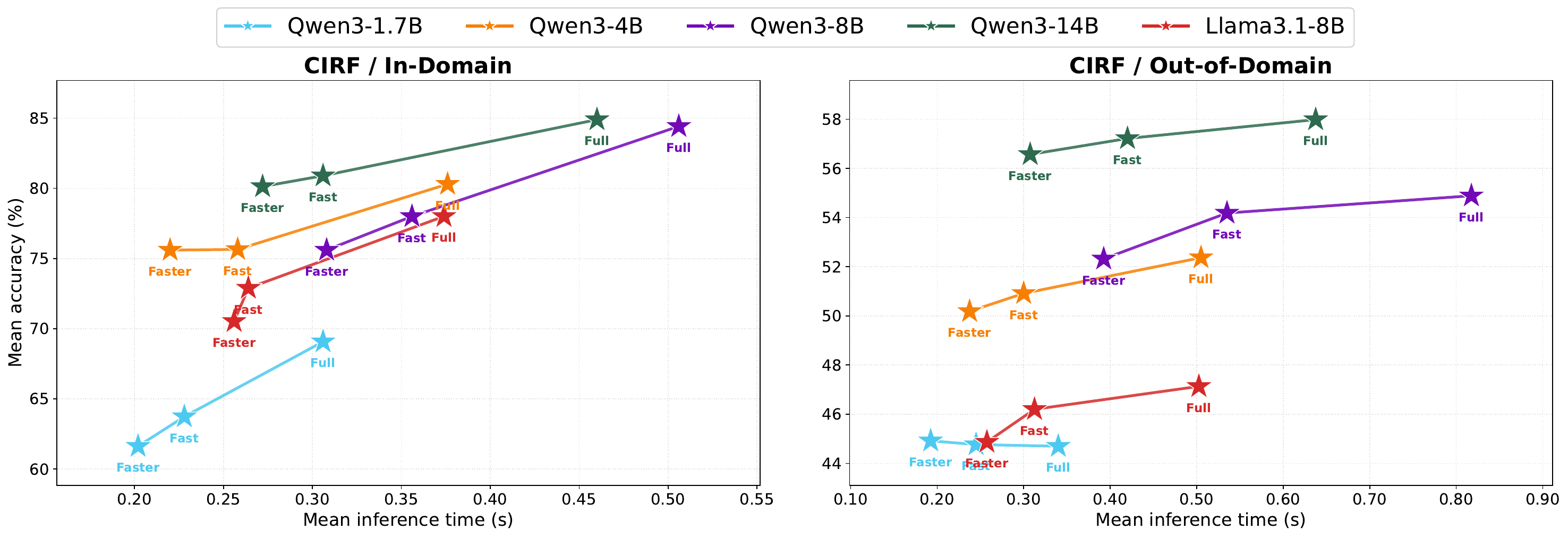}
    \caption{Accuracy vs. inference time for CIRF variants across five backbone LLMs.}
    \label{fig:cirf_comparison}
\end{figure*}

%% file: table/hyperparams.tex
\begin{table}[t!]
\centering
\small
\setlength{\tabcolsep}{4pt}
\begin{tabular}{lcccccccccc}
\toprule
\textbf{Parameter} & \textbf{Value} \\
\midrule
\multicolumn{2}{l}{\textit{\textbf{Step 1: Teacher CoT generation}}} \\
Generation temperature & 0.7 \\
Generation top-$p$ & 0.95 \\
Max generated tokens & 4096 \\

\multicolumn{2}{l}{\textit{\textbf{Step 2: Result alignment}}} \\
Generation temperature & 0.2 \\
Generation top-$p$ & 0.9 \\
Max generated tokens & 2048 \\

\multicolumn{2}{l}{\textit{\textbf{Step 3: Codebook}}} \\
Embedding model & \texttt{Qwen3-Embedding-8B} \\
Codebook vector dimension & 4096 \\
MLP hidden dimension & 1024 \\
Autoencoder training epochs & 30 \\
VQ epochs & 10 \\
Codebook batch size & 128 \\
Codebook learning rate & $10^{-4}$ \\
Gradient clip & 1.0 \\
VQ commitment $\beta$ & 1.0 \\
Sinkhorn $\lambda$ & 0.05 \\
Sinkhorn iterations & 3 \\
Codebook size $K$ & 32, 64, 128, 256 \\
Vector scale $\alpha$ & 0.01 \\

\multicolumn{2}{l}{\textit{\textbf{Step 4: LM fine-tuning}}} \\
Hidden size & 4096 \\
Layers / heads & 36 / 32 \\
Training epochs & 2 \\
Batch size & 4 \\
Gradient accumulation steps & 8 \\
Effective batch & 32 \\
Optimizer & AdamW \\
Learning rate & $2\times10^{-5}$ \\
Weight decay & 0.1 \\
Adam $\beta_1, \beta_2$ & 0.9, 0.999 \\
Adam $\epsilon$ & $10^{-8}$ \\
LR scheduler & cosine \\
Warmup steps & 100 \\
Min LR ratio & 0.1 \\
Gradient clip & 1.0 \\

\multicolumn{2}{l}{\textit{\textbf{Step 5: Evaluation}}} \\
Generation temperature & 0.7 \\
Generation top-$p$ & 0.9 \\

\multicolumn{2}{l}{\textit{\textbf{Greedy segment selection}}} \\
Loss advantage threshold $\gamma$ & 0, 0.1, 0.2 \\
\bottomrule
\end{tabular}
\caption{
Hyper-parameter settings.
}
\label{tab:hyper_parameters}
\end{table}

%% file: table/main_experiment.tex
\begin{table*}[ht]
    \centering
    \scriptsize
    \renewcommand{\arraystretch}{1.2}
    \resizebox{\textwidth}{!}{%
        \begin{tabular}{lllrrrrrrrrrrrrr}
            \toprule
            & & & \multicolumn{5}{c}{\textit{In-Domain Tasks}} & \multicolumn{4}{c}{\textit{Out-of-Domain Tasks}} \\
            \cmidrule(lr){4-8} \cmidrule(lr){9-12}
            & \textbf{Method} & \textbf{Metric} & \textbf{CoinFlip} & \textbf{Common} & \textbf{GSM8k} & \textbf{MultiArith} & \textbf{SVAMP} & \textbf{Math-500} & \textbf{BBH} & \textbf{StrategyQA} & \textbf{ScienceQA} \\
            \midrule
            \multirow{28}{*}{\rotatebox{90}{\textit{Qwen3-8B}}}
             & \multirow{2}{*}{Ans (Base)} & Acc (\%) & 34.83$\pm$1.84 & 77.67$\pm$0.85 & 17.67$\pm$1.93 & 11.11$\pm$1.81 & 29.33$\pm$0.47 & 16.40$\pm$0.85 & 37.47$\pm$0.90 & 49.13$\pm$1.05 & 75.00$\pm$0.57 \\
             & & \textcolor{gray}{\scriptsize Time(s)} & \textcolor{gray}{\scriptsize \textbf{0.26$\pm$0.00}} & \textcolor{gray}{\scriptsize \textbf{0.25$\pm$0.00}} & \textcolor{gray}{\scriptsize \textbf{0.26$\pm$0.00}} & \textcolor{gray}{\scriptsize \textbf{0.26$\pm$0.00}} & \textcolor{gray}{\scriptsize \textbf{0.26$\pm$0.00}} & \textcolor{gray}{\scriptsize \textbf{0.26$\pm$0.00}} & \textcolor{gray}{\scriptsize \textbf{0.27$\pm$0.00}} & \textcolor{gray}{\scriptsize \textbf{0.25$\pm$0.00}} & \textcolor{gray}{\scriptsize \textbf{0.26$\pm$0.00}} \\
             & \multirow{2}{*}{CoT} & Acc (\%) & \textbf{98.67$\pm$0.62} & 84.83$\pm$0.47 & \textbf{90.67$\pm$1.55} & \textbf{100.00$\pm$0.00} & \textbf{91.17$\pm$0.47} & \textbf{34.93$\pm$1.73} & \textbf{48.07$\pm$1.46} & 57.47$\pm$0.81 & 41.33$\pm$28.38 \\
             & & \textcolor{gray}{\scriptsize Time(s)} & \textcolor{gray}{\scriptsize 4.39$\pm$0.03} & \textcolor{gray}{\scriptsize 7.11$\pm$0.03} & \textcolor{gray}{\scriptsize 4.83$\pm$0.06} & \textcolor{gray}{\scriptsize 3.29$\pm$0.02} & \textcolor{gray}{\scriptsize 3.21$\pm$0.01} & \textcolor{gray}{\scriptsize 9.79$\pm$0.04} & \textcolor{gray}{\scriptsize 7.37$\pm$0.06} & \textcolor{gray}{\scriptsize 9.27$\pm$0.11} & \textcolor{gray}{\scriptsize 6.41$\pm$0.04} \\
             & \multirow{2}{*}{CoD} & Acc (\%) & 64.50$\pm$0.41 & 19.17$\pm$0.85 & 6.50$\pm$0.71 & 32.41$\pm$1.05 & 16.67$\pm$2.46 & 27.27$\pm$0.90 & 7.40$\pm$0.16 & 3.27$\pm$0.52 & 18.47$\pm$0.52 \\
             & & \textcolor{gray}{\scriptsize Time(s)} & \textcolor{gray}{\scriptsize 2.85$\pm$0.01} & \textcolor{gray}{\scriptsize 2.87$\pm$0.01} & \textcolor{gray}{\scriptsize 2.91$\pm$0.01} & \textcolor{gray}{\scriptsize 2.87$\pm$0.01} & \textcolor{gray}{\scriptsize 2.88$\pm$0.01} & \textcolor{gray}{\scriptsize 13.28$\pm$0.04} & \textcolor{gray}{\scriptsize 2.86$\pm$0.01} & \textcolor{gray}{\scriptsize 2.84$\pm$0.01} & \textcolor{gray}{\scriptsize 2.84$\pm$0.01} \\
             & \multirow{2}{*}{TokenSkip} & Acc (\%) & 53.00$\pm$20.36 & 23.50$\pm$3.54 & 36.00$\pm$0.41 & 14.07$\pm$1.05 & 45.50$\pm$2.45 & 20.80$\pm$1.14 & 14.33$\pm$1.51 & 27.33$\pm$2.78 & 50.13$\pm$4.06 \\
             & & \textcolor{gray}{\scriptsize Time(s)} & \textcolor{gray}{\scriptsize 2.89$\pm$0.42} & \textcolor{gray}{\scriptsize 2.88$\pm$0.43} & \textcolor{gray}{\scriptsize 2.89$\pm$0.43} & \textcolor{gray}{\scriptsize 2.88$\pm$0.43} & \textcolor{gray}{\scriptsize 2.89$\pm$0.43} & \textcolor{gray}{\scriptsize 14.88$\pm$2.15} & \textcolor{gray}{\scriptsize 2.91$\pm$0.42} & \textcolor{gray}{\scriptsize 3.18$\pm$0.44} & \textcolor{gray}{\scriptsize 3.18$\pm$0.43} \\
             & \multirow{2}{*}{Pause} & Acc (\%) & 74.33$\pm$7.19 & 71.83$\pm$3.30 & 32.33$\pm$2.59 & 87.96$\pm$2.88 & 67.50$\pm$4.30 & 22.33$\pm$3.20 & 45.07$\pm$0.19 & 40.00$\pm$8.74 & 70.60$\pm$1.56 \\
             & & \textcolor{gray}{\scriptsize Time(s)} & \textcolor{gray}{\scriptsize 0.74$\pm$0.02} & \textcolor{gray}{\scriptsize 0.78$\pm$0.00} & \textcolor{gray}{\scriptsize 0.80$\pm$0.00} & \textcolor{gray}{\scriptsize 0.79$\pm$0.00} & \textcolor{gray}{\scriptsize 0.79$\pm$0.00} & \textcolor{gray}{\scriptsize 12.25$\pm$0.21} & \textcolor{gray}{\scriptsize 0.79$\pm$0.00} & \textcolor{gray}{\scriptsize 0.53$\pm$0.07} & \textcolor{gray}{\scriptsize 0.69$\pm$0.02} \\
             & \multirow{2}{*}{iCoT-SI} & Acc (\%) & 25.00$\pm$5.31 & 78.17$\pm$5.04 & 24.67$\pm$1.55 & 69.44$\pm$9.26 & 60.00$\pm$7.38 & 12.53$\pm$2.41 & 40.73$\pm$2.36 & 7.40$\pm$2.12 & 73.07$\pm$0.34 \\
             & & \textcolor{gray}{\scriptsize Time(s)} & \textcolor{gray}{\scriptsize 0.73$\pm$0.25} & \textcolor{gray}{\scriptsize 0.71$\pm$0.12} & \textcolor{gray}{\scriptsize 0.79$\pm$0.18} & \textcolor{gray}{\scriptsize 0.65$\pm$0.29} & \textcolor{gray}{\scriptsize 0.69$\pm$0.24} & \textcolor{gray}{\scriptsize 11.86$\pm$0.86} & \textcolor{gray}{\scriptsize 0.69$\pm$0.14} & \textcolor{gray}{\scriptsize 0.75$\pm$0.05} & \textcolor{gray}{\scriptsize 0.63$\pm$0.12} \\
             & \multirow{2}{*}{Coconut} & Acc (\%) & 48.33$\pm$3.57 & 57.17$\pm$7.26 & 27.50$\pm$2.27 & 67.59$\pm$0.52 & 62.33$\pm$1.25 & 27.87$\pm$2.32 & 11.07$\pm$2.13 & 3.60$\pm$1.02 & 51.60$\pm$4.38 \\
             & & \textcolor{gray}{\scriptsize Time(s)} & \textcolor{gray}{\scriptsize 0.82$\pm$0.00} & \textcolor{gray}{\scriptsize 0.82$\pm$0.00} & \textcolor{gray}{\scriptsize 0.83$\pm$0.00} & \textcolor{gray}{\scriptsize 0.83$\pm$0.00} & \textcolor{gray}{\scriptsize 0.83$\pm$0.00} & \textcolor{gray}{\scriptsize 11.91$\pm$0.12} & \textcolor{gray}{\scriptsize 0.83$\pm$0.00} & \textcolor{gray}{\scriptsize 0.80$\pm$0.01} & \textcolor{gray}{\scriptsize 0.79$\pm$0.00} \\
             & \multirow{2}{*}{CODI} & Acc (\%) & 72.33$\pm$13.66 & 82.00$\pm$2.86 & 33.00$\pm$5.61 & 82.04$\pm$7.55 & 73.33$\pm$2.49 & 18.07$\pm$2.12 & 46.80$\pm$1.47 & 64.53$\pm$3.43 & 74.13$\pm$0.66 \\
             & & \textcolor{gray}{\scriptsize Time(s)} & \textcolor{gray}{\scriptsize 0.45$\pm$0.00} & \textcolor{gray}{\scriptsize 0.45$\pm$0.00} & \textcolor{gray}{\scriptsize 0.50$\pm$0.01} & \textcolor{gray}{\scriptsize 0.47$\pm$0.00} & \textcolor{gray}{\scriptsize 0.49$\pm$0.00} & \textcolor{gray}{\scriptsize 0.54$\pm$0.03} & \textcolor{gray}{\scriptsize 0.47$\pm$0.02} & \textcolor{gray}{\scriptsize 0.46$\pm$0.02} & \textcolor{gray}{\scriptsize 0.45$\pm$0.00} \\
             & \multirow{2}{*}{SIM-CoT} & Acc (\%) & 52.50$\pm$5.02 & 54.83$\pm$10.27 & 25.67$\pm$3.79 & 67.22$\pm$3.60 & 58.67$\pm$5.20 & 28.93$\pm$2.45 & 11.47$\pm$4.22 & 4.13$\pm$0.25 & 49.93$\pm$8.07 \\
             & & \textcolor{gray}{\scriptsize Time(s)} & \textcolor{gray}{\scriptsize 1.13$\pm$0.02} & \textcolor{gray}{\scriptsize 1.14$\pm$0.01} & \textcolor{gray}{\scriptsize 1.15$\pm$0.02} & \textcolor{gray}{\scriptsize 1.14$\pm$0.01} & \textcolor{gray}{\scriptsize 1.14$\pm$0.01} & \textcolor{gray}{\scriptsize 16.39$\pm$0.11} & \textcolor{gray}{\scriptsize 1.14$\pm$0.01} & \textcolor{gray}{\scriptsize 1.10$\pm$0.01} & \textcolor{gray}{\scriptsize 1.08$\pm$0.01} \\
             & \multirow{2}{*}{SoftCoT} & Acc (\%) & 60.17$\pm$2.25 & 58.33$\pm$6.06 & 21.50$\pm$2.48 & 64.44$\pm$6.71 & 53.17$\pm$1.84 & 13.67$\pm$3.67 & 30.60$\pm$5.52 & 7.07$\pm$2.68 & 62.07$\pm$5.31 \\
             & & \textcolor{gray}{\scriptsize Time(s)} & \textcolor{gray}{\scriptsize 0.87$\pm$0.01} & \textcolor{gray}{\scriptsize 0.86$\pm$0.00} & \textcolor{gray}{\scriptsize 0.88$\pm$0.01} & \textcolor{gray}{\scriptsize 0.86$\pm$0.00} & \textcolor{gray}{\scriptsize 0.87$\pm$0.01} & \textcolor{gray}{\scriptsize 13.75$\pm$0.21} & \textcolor{gray}{\scriptsize 0.87$\pm$0.01} & \textcolor{gray}{\scriptsize 0.85$\pm$0.00} & \textcolor{gray}{\scriptsize 0.83$\pm$0.02} \\
             & \multirow{2}{*}{SemCoT} & Acc (\%) & 46.33$\pm$3.70 & \textbf{98.17$\pm$0.85} & 30.33$\pm$0.47 & 80.37$\pm$2.92 & 71.00$\pm$2.04 & 19.40$\pm$8.56 & 20.93$\pm$1.79 & 3.87$\pm$1.57 & 61.33$\pm$4.34 \\
             & & \textcolor{gray}{\scriptsize Time(s)} & \textcolor{gray}{\scriptsize 0.99$\pm$0.00} & \textcolor{gray}{\scriptsize 0.99$\pm$0.00} & \textcolor{gray}{\scriptsize 1.01$\pm$0.00} & \textcolor{gray}{\scriptsize 0.99$\pm$0.00} & \textcolor{gray}{\scriptsize 0.99$\pm$0.00} & \textcolor{gray}{\scriptsize 13.58$\pm$0.63} & \textcolor{gray}{\scriptsize 0.99$\pm$0.00} & \textcolor{gray}{\scriptsize 0.96$\pm$0.01} & \textcolor{gray}{\scriptsize 0.95$\pm$0.01} \\
             \rowcolor{skyblue} & CIRF$_{\textrm{Full}}$ & Acc (\%) & 83.17$\pm$1.55 & 87.00$\pm$0.41 & 71.17$\pm$1.70 & 97.41$\pm$0.52 & 83.33$\pm$1.03 & 28.27$\pm$0.84 & 46.67$\pm$1.09 & \textbf{69.07$\pm$0.34} & 75.53$\pm$1.09 \\
             \rowcolor{skyblue} & & \textcolor{gray}{\scriptsize Time(s)} & \textcolor{gray}{\scriptsize 0.48$\pm$0.01} & \textcolor{gray}{\scriptsize 0.30$\pm$0.00} & \textcolor{gray}{\scriptsize 0.84$\pm$0.01} & \textcolor{gray}{\scriptsize 0.45$\pm$0.01} & \textcolor{gray}{\scriptsize 0.46$\pm$0.00} & \textcolor{gray}{\scriptsize 1.48$\pm$0.01} & \textcolor{gray}{\scriptsize 0.87$\pm$0.02} & \textcolor{gray}{\scriptsize 0.49$\pm$0.01} & \textcolor{gray}{\scriptsize 0.43$\pm$0.01} \\
             \rowcolor{skyblue} & CIRF$_{\textrm{Fast}}$ & Acc (\%) & 62.33$\pm$0.24 & 86.83$\pm$0.85 & 64.67$\pm$0.24 & 93.33$\pm$0.00 & 82.83$\pm$0.24 & 27.60$\pm$0.59 & 44.93$\pm$0.82 & 67.73$\pm$0.57 & \textbf{76.47$\pm$1.11} \\
             \rowcolor{skyblue} & & \textcolor{gray}{\scriptsize Time(s)} & \textcolor{gray}{\scriptsize 0.22$\pm$0.00} & \textcolor{gray}{\scriptsize 0.28$\pm$0.00} & \textcolor{gray}{\scriptsize 0.65$\pm$0.02} & \textcolor{gray}{\scriptsize 0.30$\pm$0.01} & \textcolor{gray}{\scriptsize 0.33$\pm$0.01} & \textcolor{gray}{\scriptsize 0.92$\pm$0.02} & \textcolor{gray}{\scriptsize 0.55$\pm$0.01} & \textcolor{gray}{\scriptsize 0.36$\pm$0.01} & \textcolor{gray}{\scriptsize 0.31$\pm$0.00} \\
             \rowcolor{skyblue} & CIRF$_{\textrm{Faster}}$ & Acc (\%) & 62.50$\pm$0.00 & 87.00$\pm$0.41 & 54.67$\pm$1.31 & 93.33$\pm$0.79 & 80.50$\pm$0.00 & 23.87$\pm$0.75 & 44.60$\pm$0.43 & 65.67$\pm$0.96 & 75.13$\pm$1.09 \\
             \rowcolor{skyblue} & & \textcolor{gray}{\scriptsize Time(s)} & \textcolor{gray}{\scriptsize 0.22$\pm$0.00} & \textcolor{gray}{\scriptsize 0.28$\pm$0.00} & \textcolor{gray}{\scriptsize 0.46$\pm$0.02} & \textcolor{gray}{\scriptsize 0.27$\pm$0.00} & \textcolor{gray}{\scriptsize 0.31$\pm$0.00} & \textcolor{gray}{\scriptsize 0.58$\pm$0.00} & \textcolor{gray}{\scriptsize 0.43$\pm$0.01} & \textcolor{gray}{\scriptsize 0.28$\pm$0.00} & \textcolor{gray}{\scriptsize 0.28$\pm$0.00} \\
            \midrule
            \multirow{28}{*}{\rotatebox{90}{\textit{Llama3.1-8B}}}
             & \multirow{2}{*}{Ans (Base)} & Acc (\%) & 38.50$\pm$2.68 & 73.67$\pm$1.55 & 12.50$\pm$1.78 & 27.22$\pm$0.45 & 54.00$\pm$0.71 & 9.27$\pm$0.74 & 38.47$\pm$0.66 & 61.07$\pm$1.76 & \textbf{77.13$\pm$2.04} \\
             & & \textcolor{gray}{\scriptsize Time(s)} & \textcolor{gray}{\scriptsize \textbf{0.24$\pm$0.00}} & \textcolor{gray}{\scriptsize \textbf{0.24$\pm$0.00}} & \textcolor{gray}{\scriptsize \textbf{0.25$\pm$0.00}} & \textcolor{gray}{\scriptsize \textbf{0.24$\pm$0.00}} & \textcolor{gray}{\scriptsize \textbf{0.24$\pm$0.00}} & \textcolor{gray}{\scriptsize \textbf{0.25$\pm$0.00}} & \textcolor{gray}{\scriptsize \textbf{0.26$\pm$0.00}} & \textcolor{gray}{\scriptsize \textbf{0.24$\pm$0.00}} & \textcolor{gray}{\scriptsize \textbf{0.24$\pm$0.00}} \\
             & \multirow{2}{*}{CoT} & Acc (\%) & \textbf{98.17$\pm$0.62} & 81.83$\pm$1.25 & \textbf{81.17$\pm$2.09} & \textbf{98.70$\pm$0.52} & \textbf{86.50$\pm$0.41} & 24.93$\pm$1.20 & \textbf{47.13$\pm$1.09} & \textbf{69.13$\pm$2.00} & 69.13$\pm$0.34 \\
             & & \textcolor{gray}{\scriptsize Time(s)} & \textcolor{gray}{\scriptsize 4.58$\pm$0.08} & \textcolor{gray}{\scriptsize 6.37$\pm$0.03} & \textcolor{gray}{\scriptsize 4.45$\pm$0.03} & \textcolor{gray}{\scriptsize 3.06$\pm$0.02} & \textcolor{gray}{\scriptsize 3.03$\pm$0.02} & \textcolor{gray}{\scriptsize 8.72$\pm$0.04} & \textcolor{gray}{\scriptsize 6.32$\pm$0.04} & \textcolor{gray}{\scriptsize 7.45$\pm$0.10} & \textcolor{gray}{\scriptsize 6.06$\pm$0.04} \\
             & \multirow{2}{*}{CoD} & Acc (\%) & 93.00$\pm$0.41 & 70.67$\pm$1.25 & 58.00$\pm$0.82 & 93.52$\pm$0.69 & 75.17$\pm$1.55 & \textbf{30.53$\pm$0.96} & 41.87$\pm$0.98 & 67.20$\pm$0.71 & 75.00$\pm$0.00 \\
             & & \textcolor{gray}{\scriptsize Time(s)} & \textcolor{gray}{\scriptsize 0.48$\pm$0.00} & \textcolor{gray}{\scriptsize 0.46$\pm$0.01} & \textcolor{gray}{\scriptsize 1.86$\pm$0.03} & \textcolor{gray}{\scriptsize 0.93$\pm$0.03} & \textcolor{gray}{\scriptsize 1.25$\pm$0.04} & \textcolor{gray}{\scriptsize 5.82$\pm$0.17} & \textcolor{gray}{\scriptsize 1.38$\pm$0.02} & \textcolor{gray}{\scriptsize 0.84$\pm$0.02} & \textcolor{gray}{\scriptsize 0.46$\pm$0.01} \\
             & \multirow{2}{*}{TokenSkip} & Acc (\%) & 87.50$\pm$4.81 & 23.67$\pm$2.36 & 31.67$\pm$2.72 & 11.11$\pm$3.95 & 64.83$\pm$1.55 & 10.80$\pm$1.85 & 19.07$\pm$1.84 & 31.00$\pm$8.09 & 46.87$\pm$3.03 \\
             & & \textcolor{gray}{\scriptsize Time(s)} & \textcolor{gray}{\scriptsize 1.89$\pm$0.04} & \textcolor{gray}{\scriptsize 2.02$\pm$0.00} & \textcolor{gray}{\scriptsize 1.93$\pm$0.14} & \textcolor{gray}{\scriptsize 2.02$\pm$0.01} & \textcolor{gray}{\scriptsize 2.04$\pm$0.03} & \textcolor{gray}{\scriptsize 9.34$\pm$1.42} & \textcolor{gray}{\scriptsize 2.00$\pm$0.04} & \textcolor{gray}{\scriptsize 1.89$\pm$0.11} & \textcolor{gray}{\scriptsize 1.98$\pm$0.04} \\
             & \multirow{2}{*}{Pause} & Acc (\%) & 57.50$\pm$4.26 & 70.17$\pm$3.68 & 15.00$\pm$2.83 & 81.67$\pm$6.35 & 71.50$\pm$2.86 & 9.67$\pm$1.48 & 36.40$\pm$3.05 & 37.40$\pm$15.29 & 69.47$\pm$3.87 \\
             & & \textcolor{gray}{\scriptsize Time(s)} & \textcolor{gray}{\scriptsize 0.15$\pm$0.07} & \textcolor{gray}{\scriptsize 0.48$\pm$0.17} & \textcolor{gray}{\scriptsize 0.18$\pm$0.08} & \textcolor{gray}{\scriptsize 0.13$\pm$0.08} & \textcolor{gray}{\scriptsize 0.11$\pm$0.05} & \textcolor{gray}{\scriptsize 0.98$\pm$0.77} & \textcolor{gray}{\scriptsize 0.45$\pm$0.21} & \textcolor{gray}{\scriptsize 0.46$\pm$0.18} & \textcolor{gray}{\scriptsize 0.44$\pm$0.18} \\
             & \multirow{2}{*}{iCoT-SI} & Acc (\%) & 36.00$\pm$7.12 & 79.50$\pm$0.71 & 15.83$\pm$4.11 & 57.22$\pm$7.49 & 72.67$\pm$1.65 & 11.07$\pm$0.74 & 32.07$\pm$4.49 & 49.00$\pm$4.60 & 73.73$\pm$2.00 \\
             & & \textcolor{gray}{\scriptsize Time(s)} & \textcolor{gray}{\scriptsize 0.22$\pm$0.09} & \textcolor{gray}{\scriptsize 0.29$\pm$0.11} & \textcolor{gray}{\scriptsize 0.22$\pm$0.08} & \textcolor{gray}{\scriptsize 0.20$\pm$0.08} & \textcolor{gray}{\scriptsize 0.20$\pm$0.09} & \textcolor{gray}{\scriptsize 0.97$\pm$0.73} & \textcolor{gray}{\scriptsize 0.37$\pm$0.12} & \textcolor{gray}{\scriptsize 0.32$\pm$0.14} & \textcolor{gray}{\scriptsize 0.37$\pm$0.12} \\
             & \multirow{2}{*}{Coconut} & Acc (\%) & 37.33$\pm$3.47 & 64.83$\pm$3.09 & 11.50$\pm$3.34 & 35.00$\pm$2.72 & 56.17$\pm$0.62 & 10.47$\pm$1.36 & 25.87$\pm$0.82 & 16.27$\pm$3.70 & 66.33$\pm$0.34 \\
             & & \textcolor{gray}{\scriptsize Time(s)} & \textcolor{gray}{\scriptsize 0.33$\pm$0.05} & \textcolor{gray}{\scriptsize 0.52$\pm$0.06} & \textcolor{gray}{\scriptsize 0.58$\pm$0.09} & \textcolor{gray}{\scriptsize 0.55$\pm$0.04} & \textcolor{gray}{\scriptsize 0.41$\pm$0.05} & \textcolor{gray}{\scriptsize 3.63$\pm$0.88} & \textcolor{gray}{\scriptsize 0.53$\pm$0.02} & \textcolor{gray}{\scriptsize 0.50$\pm$0.02} & \textcolor{gray}{\scriptsize 0.44$\pm$0.04} \\
             & \multirow{2}{*}{CODI} & Acc (\%) & 63.00$\pm$6.68 & 74.50$\pm$2.94 & 8.50$\pm$2.16 & 49.26$\pm$8.92 & 59.00$\pm$6.16 & 7.20$\pm$0.28 & 33.93$\pm$0.50 & 53.87$\pm$6.63 & 68.00$\pm$1.07 \\
             & & \textcolor{gray}{\scriptsize Time(s)} & \textcolor{gray}{\scriptsize 0.37$\pm$0.00} & \textcolor{gray}{\scriptsize 0.37$\pm$0.00} & \textcolor{gray}{\scriptsize 0.37$\pm$0.00} & \textcolor{gray}{\scriptsize 0.37$\pm$0.00} & \textcolor{gray}{\scriptsize 0.37$\pm$0.00} & \textcolor{gray}{\scriptsize 0.38$\pm$0.00} & \textcolor{gray}{\scriptsize 0.37$\pm$0.00} & \textcolor{gray}{\scriptsize 0.37$\pm$0.00} & \textcolor{gray}{\scriptsize 0.37$\pm$0.00} \\
             & \multirow{2}{*}{SIM-CoT} & Acc (\%) & 28.67$\pm$2.01 & 60.67$\pm$5.39 & 11.17$\pm$1.31 & 35.00$\pm$4.03 & 55.17$\pm$2.72 & 10.60$\pm$0.16 & 23.93$\pm$4.16 & 24.20$\pm$3.74 & 59.07$\pm$5.73 \\
             & & \textcolor{gray}{\scriptsize Time(s)} & \textcolor{gray}{\scriptsize 0.24$\pm$0.03} & \textcolor{gray}{\scriptsize 0.52$\pm$0.05} & \textcolor{gray}{\scriptsize 0.48$\pm$0.04} & \textcolor{gray}{\scriptsize 0.46$\pm$0.05} & \textcolor{gray}{\scriptsize 0.37$\pm$0.04} & \textcolor{gray}{\scriptsize 3.04$\pm$0.74} & \textcolor{gray}{\scriptsize 0.51$\pm$0.03} & \textcolor{gray}{\scriptsize 0.48$\pm$0.01} & \textcolor{gray}{\scriptsize 0.44$\pm$0.05} \\
             & \multirow{2}{*}{SoftCoT} & Acc (\%) & 46.67$\pm$5.17 & 56.00$\pm$4.60 & 9.50$\pm$1.08 & 28.15$\pm$3.02 & 47.50$\pm$4.97 & 8.80$\pm$4.97 & 20.27$\pm$4.43 & 17.60$\pm$2.57 & 53.87$\pm$8.88 \\
             & & \textcolor{gray}{\scriptsize Time(s)} & \textcolor{gray}{\scriptsize 0.50$\pm$0.07} & \textcolor{gray}{\scriptsize 0.71$\pm$0.01} & \textcolor{gray}{\scriptsize 0.72$\pm$0.01} & \textcolor{gray}{\scriptsize 0.70$\pm$0.01} & \textcolor{gray}{\scriptsize 0.58$\pm$0.09} & \textcolor{gray}{\scriptsize 2.92$\pm$3.26} & \textcolor{gray}{\scriptsize 0.69$\pm$0.01} & \textcolor{gray}{\scriptsize 0.63$\pm$0.02} & \textcolor{gray}{\scriptsize 0.66$\pm$0.00} \\
             & \multirow{2}{*}{SemCoT} & Acc (\%) & 55.67$\pm$15.34 & 71.50$\pm$19.80 & 11.83$\pm$7.19 & 42.04$\pm$29.99 & 49.17$\pm$12.98 & 19.60$\pm$4.42 & 41.27$\pm$23.15 & 20.47$\pm$7.91 & 70.27$\pm$14.97 \\
             & & \textcolor{gray}{\scriptsize Time(s)} & \textcolor{gray}{\scriptsize 0.65$\pm$0.16} & \textcolor{gray}{\scriptsize 0.84$\pm$0.01} & \textcolor{gray}{\scriptsize 0.87$\pm$0.01} & \textcolor{gray}{\scriptsize 0.84$\pm$0.01} & \textcolor{gray}{\scriptsize 0.78$\pm$0.02} & \textcolor{gray}{\scriptsize 5.41$\pm$3.28} & \textcolor{gray}{\scriptsize 0.84$\pm$0.01} & \textcolor{gray}{\scriptsize 0.80$\pm$0.04} & \textcolor{gray}{\scriptsize 0.74$\pm$0.06} \\
             \rowcolor{skyblue} & CIRF$_{\textrm{Full}}$ & Acc (\%) & 77.00$\pm$0.82 & 81.83$\pm$0.62 & 56.17$\pm$2.59 & \textbf{98.52$\pm$0.69} & 76.50$\pm$0.71 & 11.27$\pm$0.09 & 34.13$\pm$0.96 & 68.93$\pm$0.82 & 74.20$\pm$0.71 \\
             \rowcolor{skyblue} & & \textcolor{gray}{\scriptsize Time(s)} & \textcolor{gray}{\scriptsize 0.34$\pm$0.01} & \textcolor{gray}{\scriptsize 0.25$\pm$0.00} & \textcolor{gray}{\scriptsize 0.57$\pm$0.01} & \textcolor{gray}{\scriptsize 0.38$\pm$0.00} & \textcolor{gray}{\scriptsize 0.33$\pm$0.01} & \textcolor{gray}{\scriptsize 0.76$\pm$0.03} & \textcolor{gray}{\scriptsize 0.48$\pm$0.01} & \textcolor{gray}{\scriptsize 0.44$\pm$0.01} & \textcolor{gray}{\scriptsize 0.33$\pm$0.00} \\
             \rowcolor{skyblue} & CIRF$_{\textrm{Fast}}$ & Acc (\%) & 57.17$\pm$0.24 & 81.17$\pm$1.43 & 51.83$\pm$1.43 & 96.11$\pm$0.79 & \textbf{78.17$\pm$1.03} & 9.33$\pm$0.25 & 35.27$\pm$1.36 & 64.47$\pm$0.52 & 75.73$\pm$1.00 \\
             \rowcolor{skyblue} & & \textcolor{gray}{\scriptsize Time(s)} & \textcolor{gray}{\scriptsize 0.16$\pm$0.00} & \textcolor{gray}{\scriptsize 0.19$\pm$0.00} & \textcolor{gray}{\scriptsize 0.45$\pm$0.01} & \textcolor{gray}{\scriptsize 0.29$\pm$0.00} & \textcolor{gray}{\scriptsize 0.23$\pm$0.00} & \textcolor{gray}{\scriptsize 0.44$\pm$0.01} & \textcolor{gray}{\scriptsize 0.30$\pm$0.00} & \textcolor{gray}{\scriptsize 0.30$\pm$0.01} & \textcolor{gray}{\scriptsize 0.21$\pm$0.00} \\
             \rowcolor{skyblue} & CIRF$_{\textrm{Faster}}$ & Acc (\%) & 57.50$\pm$0.00 & 82.67$\pm$0.62 & 45.33$\pm$0.85 & 90.93$\pm$0.69 & 76.17$\pm$0.47 & 9.27$\pm$0.38 & 34.47$\pm$0.41 & 60.53$\pm$0.68 & 75.20$\pm$0.49 \\
             \rowcolor{skyblue} & & \textcolor{gray}{\scriptsize Time(s)} & \textcolor{gray}{\scriptsize 0.17$\pm$0.00} & \textcolor{gray}{\scriptsize 0.21$\pm$0.00} & \textcolor{gray}{\scriptsize 0.39$\pm$0.01} & \textcolor{gray}{\scriptsize 0.28$\pm$0.00} & \textcolor{gray}{\scriptsize 0.23$\pm$0.00} & \textcolor{gray}{\scriptsize 0.32$\pm$0.01} & \textcolor{gray}{\scriptsize 0.24$\pm$0.00} & \textcolor{gray}{\scriptsize 0.27$\pm$0.01} & \textcolor{gray}{\scriptsize 0.20$\pm$0.00} \\
            \bottomrule
        \end{tabular}}
    \caption{Accuracy (\%) and inference time (seconds) across baselines and CIRF variants (\textbf{Full}: $\gamma=0.0$, \textbf{Fast}: $\gamma=0.1$, \textbf{Faster}: $\gamma=0.2$). Temperature=0.7, mean$\pm$std over 3 runs. \textbf{Bold} indicates the best result per column within each model.}
    \label{tab:main-results}
\end{table*}